\documentclass[sigconf]{acmart}%

\usepackage[english]{babel}
\usepackage{hyphenat}

\usepackage{graphicx}
\graphicspath{{./figs/}}
\usepackage{subcaption}

\usepackage[ruled,linesnumbered]{algorithm2e}
\usepackage{amsmath}

\usepackage{amssymb}

\usepackage{multirow}

\newtheorem{define}{Definition}[section]

\AtBeginDocument{%
  \providecommand\BibTeX{{%
    \normalfont B\kern-0.5em{\scshape i\kern-0.25em b}\kern-0.8em\TeX}}}

\setcopyright{acmcopyright}
\copyrightyear{2020}
\acmYear{2020}
\acmDOI{12.1234/1234567.1234567}

\acmConference[WWW 2021]{WWW 2021: 30th The Web Conference}{April 2021}{Ljubljana, Slovenia}
\acmBooktitle{WWW 2021: 30th The Web Conference, April 2021, Ljubljana, Slovenia}
\acmPrice{00.00}
\acmISBN{978-1-4503-XXXX-X/20/08}

\begin{document}

\title[JITuNE: Just-In-Time NE Hyperparameter Tuning]{JITuNE: Just-In-Time Hyperparameter Tuning for\\Network Embedding Algorithms}

\author{$^1$Mengying Guo, $^2$Tao Yi}
\affiliation{%
  \institution{$^1$ICT, CAS\qquad$^2$Tsinghua University}
  \city{Beijing}
  \country{China}
}

\author{Yuqing Zhu}
\authornote{Contact: zhuyuqing$@$tsinghua.edu.cn}
\affiliation{%
  \institution{Tsinghua University}
  \city{Beijing}
  \country{China}
}

\author{Yungang Bao}
\affiliation{%
  \institution{ICT, CAS}
  \city{Beijing}
  \country{China}
}


\begin{abstract}
Network embedding (NE) can generate succinct node representations for massive-scale networks and enable direct applications of common machine learning methods to the network structure. Various NE algorithms have been proposed and used in a number of applications, such as node classification and link prediction. NE algorithms typically contain hyperparameters that are key to performance, but the hyperparameter tuning process can be time consuming. It is desirable to have the hyperparameters tuned within a specified length of time. Although AutoML methods have been applied to the hyperparameter tuning of NE algorithms, the problem of how to tune hyperparameters in a given period of time is not studied for NE algorithms before. In this paper, we propose \textbf{JITuNE}, a just-in-time hyperparameter tuning framework for NE algorithms. Our JITuNE framework enables the time-constrained hyperparameter tuning for NE algorithms by employing the tuning over hierarchical network synopses and transferring the knowledge obtained on synopses to the whole network. The hierarchical generation of synopsis and a time-constrained tuning method enable the constraining of overall tuning time. Extensive experiments demonstrate that JITuNE can significantly improve performances of NE algorithms, outperforming state-of-the-art methods within the same number of algorithm runs.
\end{abstract}


\maketitle

\section{Introduction}

Networks are ubiquitous in real-world applications, e.g., social networks~\cite{blog}, reference networks~\cite{arxiv,wiki}, and web networks~\cite{topcat}. Due to the massive scale of real-world networks and their sparse representation of high dimensions, it is difficult to directly apply the common machine learning methods to network applications. To enable an effective learning on massive-scale networks, many research efforts have been devoted to finding effective representations of networks, which is called network embedding~\cite{cui2018survey,zhang2018network,hamilton2017representation}. Network embedding (NE) can learn succinct representations for a network or nodes of a network. For the past few years, we have witnessed a large number of NE algorithms being proposed~\cite{arope,deepwalk,fastgcn,node2vec}.

NE algorithms typically contain a number of hyperparameters that have significant impacts on their performance. While the automatic tuning of hyperparameters has been recognized generally, existent hyperparameter tuning frameworks cannot be directly applied to NE algorithms. On the one hand, NE algorithms are computationally expensive, especially on massive networks. While hyperparameter tuning of NE algorithms is sensitive to the input networks, directly incorporating NE algorithms into general hyperparameter tuning frameworks can take a prohibitively long time. On the other hand, even if common hyperparameter tuning frameworks provide some early-stopping and space-pruning measures, the special structure of network demonstrates different behaviors in NE algorithms such that these measures are not applicable. For each network, an NE algorithm must be run many times to collect samples for hyperparameter tuning. If the network size is large, it can take a prohibitively long time for tuning. As a result, the time for hyperparameter tuning of NE algorithms is typically long.
\begin{figure*}[t]
    \centering
        \includegraphics[width=.7\textwidth]{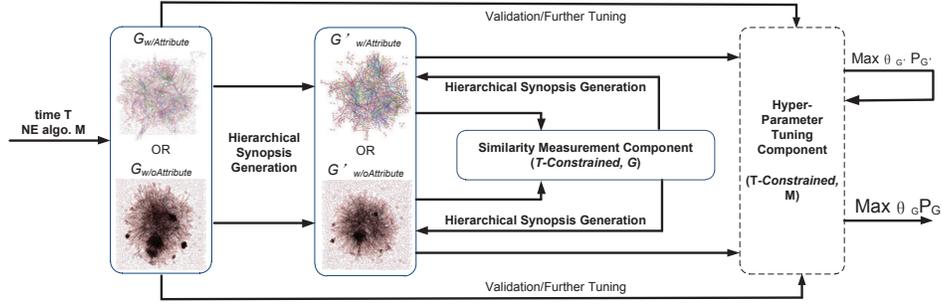}\vspace{-3pt}%
        \caption{The time-constrained tuning framework of JITuNE, returning a best hyperparameter configuration just in time: the synopsis generation module, the similarity measurement module, and the hyperparameter tuning model. $P_{G'}$ and $P_G$ is the performance achieved on G' or G using configuration $\theta_{G'}$ or $\theta_G$.}\vspace{-9pt}
        \label{fig:jitune} %
\end{figure*}

The specific requirements of automatic hyperparameter tuning for NE algorithms~\cite{autone} are noticed recently. However, the problem of prohibitively long time for hyperparameter tuning is not addressed. In real-world applications, users are not patient to wait for a long time. They are expecting to get a good result in time. Two main challenges exist in applying hyperparameter tuning of NE algorithms to real-world applications:\vspace{-6pt}
\begin{itemize}
  \item \textbf{Time-constrained algorithm runtime on massive-scale networks.} NE algorithms are generally applied to massive-scale networks. This fact indicates the running time of an NE algorithm would be long. Hyperparameter tuning would require to run an NE algorithm for many times for sample collection. We cannot constrain the time in hyperparameter tuning for an NE algorithm if the NE runtime on massive-scale networks cannot be constrained.\vspace{3pt}
  \item \textbf{Time-constrained optimization.} Hyperparameter tuning is an optimization process. Existing general hyperparameter tuning methods run progressively until decisions are made explicitly to stop the process. They leave users at wild again with another hyperparameter tuning problem, i.e., the number of samples to be collected before starting the progressive tuning process. The main problem of existing methods is that they have not designed the optimization process with regard to the time constraint. A time-constrained optimization process must in turn consider the previous challenge as well.\vspace{-9pt}
\end{itemize}

A recent work~\cite{autone} on hyperparameter tuning for NE algorithms proposes a solution based on sub-network sampling and the transfer of tuning knowledge, improving over general hyperparameter tuning methods. But it has three limitations. First, the random sampling of sub-networks can fail to represent any feature of the whole network if no prior knowledge is incorporated in the sampling. But such prior knowledge is a strong requirement for a method targeting at automating the tuning process. Second, the transferability is made through the sharing of parameters in a Gaussian process-based Bayesian optimization, thus requiring the hyperparameters under tune be numeric. But this is not always true as categorical hyperparameters also exist~\cite{gcn,fastgcn}. Third, the running time is extremely long due to sampling the multiple sub-networks and computing network similarity for transferability. The time gaining from tuning small networks is wasted in sampling and tuning multiple sub-networks, as well as executing extensive matrix computations.

In this paper, we propose \textbf{JITuNE}, a just-in-time hyperparameter tuning framework for NE algorithms. Our JITuNE framework enables the time-constrained hyperparameter tuning for NE algorithms on massive-scale networks. To constrain the runtime of NE algorithms, we propose to tune the hyperparameters of NE algorithms based on hierarchical network synopses, instead of the original network. Synopsis is an easy-to-construct representation that yields good approximations of the relevant properties of a massive dataset. As for massive network and network embedding, we require the network synopsis to be representative and small such that the hyperparameter tuning process of NE algorithms can finish in a reasonable time. The tuning of different NE algorithms can thus be constrained by using synopses at different hierarchies. Thanks to the representativeness of synopsis, the tuning knowledge can be directly transferred to the hyperparameter tuning on the original network. Besides, JITuNE has a time-constrained hyperparameter tuning process that samples in the hyperparameter space and that runs the NE algorithm in a given time budget. The hierarchical generation of synopsis and a time-constrained tuning method enables the constraining of the overall tuning time.

In sum, we make the following contributions in this paper:\vspace{-3pt}
\begin{itemize}
  \item We investigate the time-constrained hyperparameter tuning problem for NE algorithms on massive-scale networks, and propose a simple yet powerful framework of JITuNE to automate the hyperparameter tuning process.
  \item We propose the JITuNE method that can efficiently scale up to massive real-world networks and bound the tuning time, based on the hierarchical synopsis of networks and a time-constrained tuning procedure. In particular, the hierarchical synopsis generation tackles the challenge of a constrained runtime for NE algorithms, while the sampling-and-search based optimization method addresses the challenge of time-constrained optimization process.
  \item We carry out extensive experiments tuning representative NE algorithms on seven real-world networks. Experimental results demonstrate that JITuNE outperforms the state-of-the-art hyperparameter tuning methods for NE algorithms.\vspace{-3pt}
\end{itemize}

The rest of the paper is organized as follows. In Section~\ref{sec:preliminary}, we formally define the research problem. We detail the JITuNE framework in Section~\ref{sec:jitune}. In Section~\ref{sec:exp}, we report the experimental results. We summarize the related works in Section~\ref{sec:related} and conclude in Section~\ref{sec:conclude}.\vspace{-6pt}

\section{Preliminaries}
\label{sec:preliminary}%

\subsection{Definitions}

\begin{define}
\textbf{Graphs}. Networks are represented as graphs. Directed or undirected graphs are considered with or without node labels, node attributes and edge attributes. Formally, such a general graph can be represented as:
\begin{equation}
G=(V,E,L, X_V,X_E),
\end{equation}
where $V$ denotes the set of nodes, $E$ denotes the set of edges, $L$ denotes the set of node labels, $X_V$ denotes the set of node attributes,  and $X_E$ denotes the set of edge attributes. Specifically, in some cases, $L$, $X_V$ and $X_E$ might be unavailable, i.e., $L=\varnothing$, $X_V=\varnothing$, or $X_E=\varnothing$.
\end{define}

\begin{define}
\textbf{Embeddings.} In this paper, embeddings denote the $d$-dimensional learned vectors for graph nodes. Formally, let $M$ be a network embedding algorithm and $h_v$ denote the node embedding for the node $v\in V$, where $h_v=M(G,v)$.
\end{define}

That is, for a graph $G$, an NE algorithm $M$ outputs node embeddings for $G$, when given a hyperparameter configuration $\theta_G$. The dimension $d$ of the embedding space is a member of $\theta_G$, i.e., $d\in\theta_G$.

\begin{define}
\textbf{Performance.} $P_G(M,A,\theta_G)$ denotes the performance function of $M$ on the validation dataset of a downstream application $A$, e.g., node classification or link prediction, with regard to the graph $G$ and the hyperparameter configuration $\theta_G$.
\label{def:perf}
\end{define}

Note that, the function for $P_G$ is generally unknown. It can take any shape or form.

\subsection{Problem Statement}

We study the optimization problem of $P_G(M,A,\theta_G)$. Our goal is to find, in a given length of time $T$, a hyperparameter configuration $\theta_G$ that can enable an NE algorithm $M$ to achieve the best performance $P_G$ on a graph $G$. Depending on the performance metric used, the objective can be to minimize or maximize $P_G$ within the space of $\theta_G$, based on a sample set collected within time $T$.

Two subproblems exists for this optimization problem. First, as $T$ is constrained, there might not be enough samples for solving the optimization problem, if samples are collected with regard to $G$. Hence, we need to find a network $G'=(V',E')$, which captures the global and local structure of $G$, but is much smaller than the original network, i.e., $|V'|<<|V|$ and $|E'|<<|E|$. We denote $G'$ as the \emph{\textbf{synopsis}} of $G$. It is required that, the similarity of $G$ and $G'$ is larger than a threshold to enable a valid transfer of the hyperparameter tuning result from $G'$ to $G$. Then follows the second subproblem, i.e., how to optimize $P_G$ through $P_{G'}$ within $T$.

\section{The JITuNE Framework}%
\label{sec:jitune}

In this section, we describe our framework in detail. We start with the overall framework of our proposed JITuNE framework. Then, we present the details of each component in JITuNE and describe how the two subproblems of the just-in-time tuning problem are addressed.

\subsection{Framework Overview}

Illustrated in Figure~\ref{fig:jitune}, the JITuNE framework has three main components, i.e., the hierarchical synopsis generation module, the similarity measurement module, and the hyperparameter tuning module. For the tuning process, JITuNE relies on the last two modules for time constraining. The two modules decide the number of synopsis hierarchies to generate and the number of hyperparameter configurations to test (i.e., the rounds of NE algorithms to run).

 Given a network represented by $G$, an NE algorithm $M$ and a time constraint $T$ as input, JITuNE first runs $M$ once on $G$ for an estimation of one-round running time $t_G$. Then the hierarchical synopsis generation module of JITuNE generates a series of network synopses, until the similarity measurement module signals a stop. With the smallest synopsis $G'$ and $t_G$, the hyperparameter tuning component tunes the hyperparameters of $M$ on the synopsis graph $G'$. The optimal configuration of hyperparameters $\theta'$ is output and applied to the original network $G$ for validation. If extra time is given, further tuning steps can be executed on the original graph till the given time is run out of.

\subsection{Network Synopsis Generation}%
\label{sec:synopsis}

We exploit multi-level graph representation learning methods to generate hierarchical synopses for a given massive-scale network. The generation of synopsis at each hierarchy depends on the last hierarchy. For networks without attribute information, we apply the HARP method~\cite{harp}; for those networks with attribute information, we apply the H-GCN method~\cite{hgcn}. As for HARP, we only adopt the coarsening step, which has a time complexity of $\mathcal{O}(|V|)$.  Hence, the synopsis generation for graphs without attributes takes much shorter time than the NE algorithms, which tend to be in $\mathcal{O}(|E|)$~\cite{line} or higher~\cite{arope,gcn} complexities. For graphs with attributes, the adopted graph coarsening method of H-GCN is as asymptotically complex as GCN. In this case, the graph coarsening step for a hierarchy is considered as a run of the original algorithm.

\begin{figure}[t]
    \begin{subfigure}{0.2\textwidth}%
    \vspace{7pt}
    \centering\captionsetup{width=.95\textwidth}%
        \includegraphics[width=\textwidth,height=85pt]{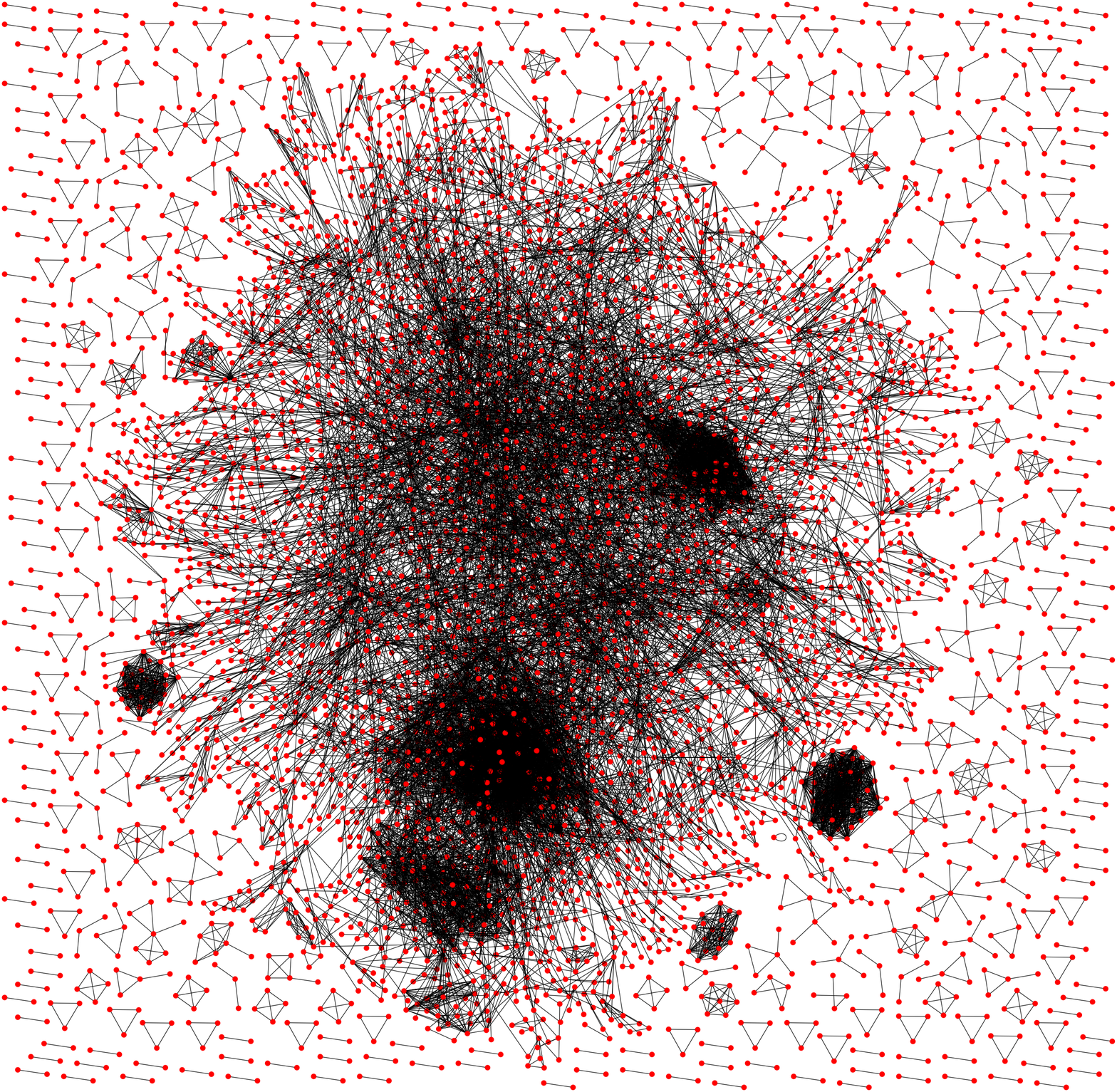}%
        \vspace{6pt}
        \caption{Original graph.}
        \vspace{-3pt}
        \label{fig:arxiv:ori} 
    \end{subfigure}
    \begin{subfigure}{0.2\textwidth}%
    \vspace{7pt}
    \centering\captionsetup{width=.95\textwidth}%
        \includegraphics[width=0.88\textwidth]{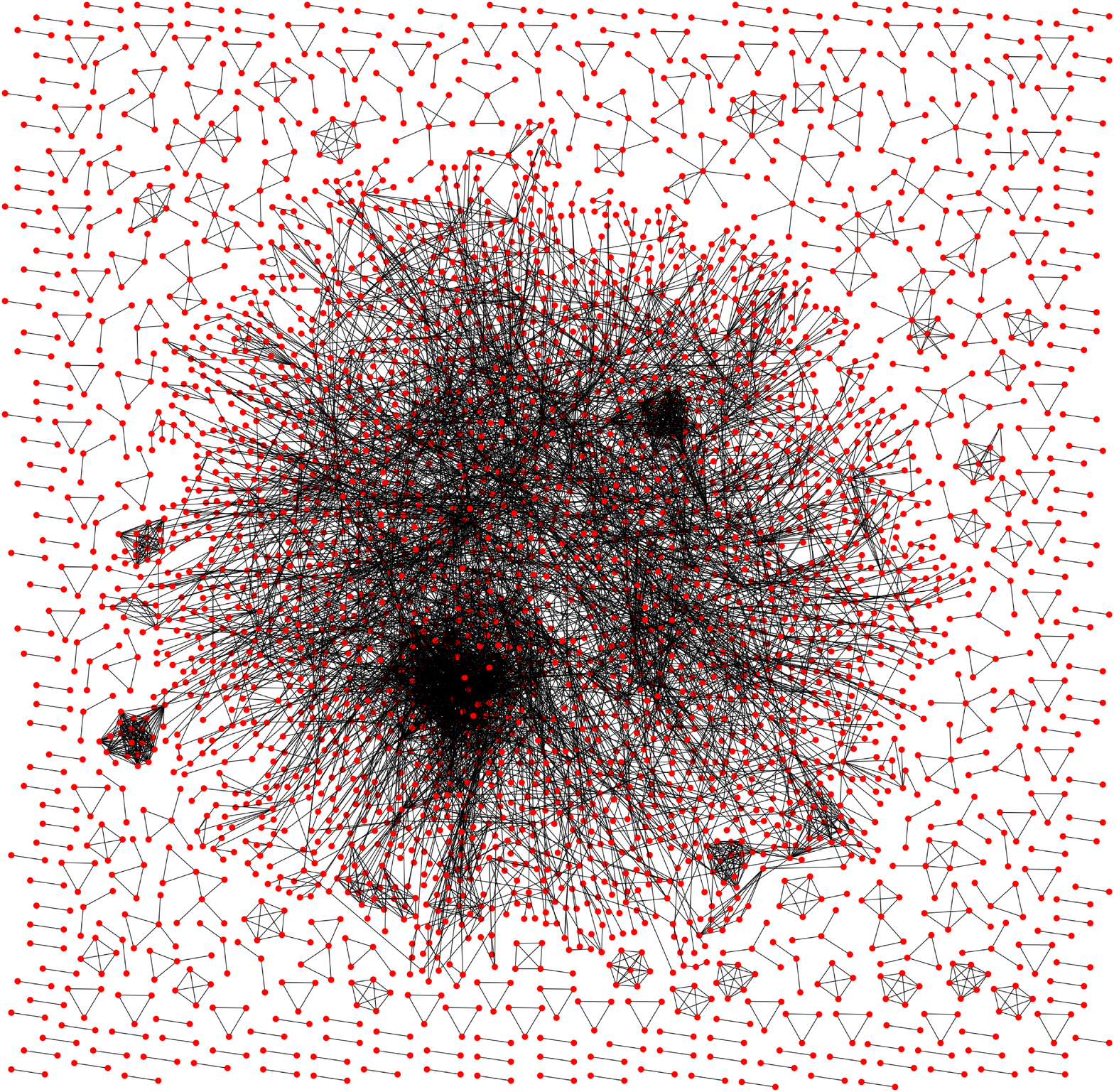}%
        \vspace{6pt}
        \caption{Network synopsis.}
        \vspace{-3pt}
        \label{fig:arxiv:syn} 
    \end{subfigure}
        \caption{The synopsis of the network Arxiv with disconnected components.}\vspace{-18pt}
    \label{fig:arxiv} 
\end{figure}

\textbf{Network without attribute information.} The network coarsening process of HARP consists of node grouping and node collapsing. On node grouping, a pair of similar nodes are grouped together. The grouped nodes are collapsed into a single node for the next hierarchy of the network synopsis. Node grouping can be edge-based or star-based. Edge-based grouping considers a pair of nodes connected by an edge, while star-based groups two nodes with the same neighbour, i.e., connected to one same node. With the HARP method, the star-based condition is first applied with the edge-based condition followed. We illustrate one synopsis of the Arxiv network in Figure~\ref{fig:arxiv}. Arxiv is a network with disconnected components. The original network is visualized in Figure~\ref{fig:arxiv:ori}, while the synopsis generated by HARP is in Figure~\ref{fig:arxiv:syn}. As the synopsis network has a much smaller size, the visualized network has a much lighter intensity.

We choose the HARP method in graph synopsis generation for two main reasons. First, the edge-based and the star-based groupings preserve the first proximity and the second proximity of  nodes respectively~\cite{harp}. This aligns the synopsis generation step with the NE algorithm tuning process, as network embedding algorithms have also the objectives of preserving the first and the second proximities of nodes. Thus, the local information and the global information such as disconnected components and communities are preserved across different phases of the tuning process. Second, the algorithm complexity of HARP is relatively low. Hence, time can be left for the tuning phase.

\begin{figure}[t]
    \begin{subfigure}{0.2\textwidth}%
    \centering\captionsetup{width=.95\textwidth}%
        \includegraphics[width=\textwidth]{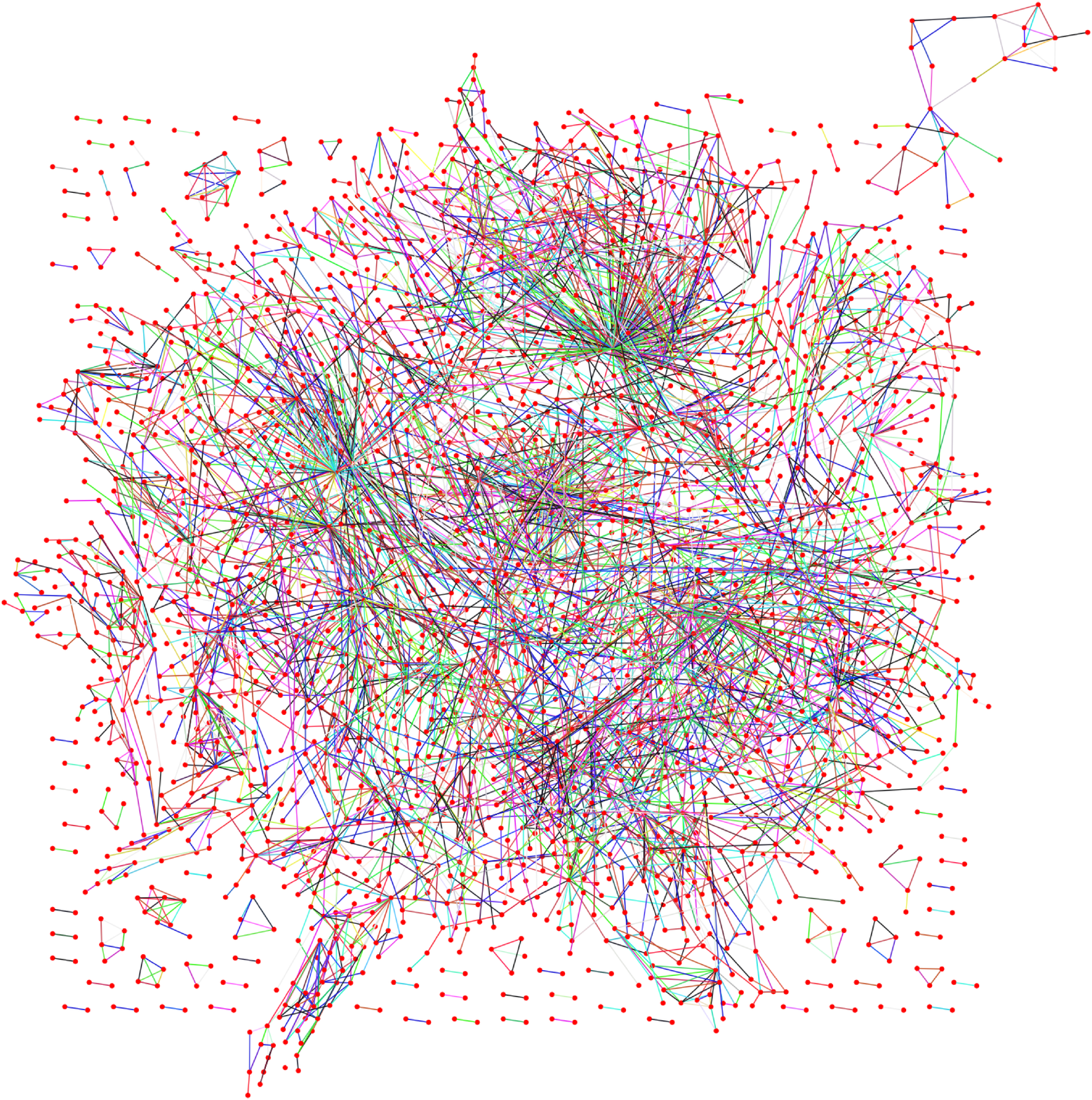}%
        \caption{Original graph.}
        \vspace{-4pt}
        \label{fig:cora:ori} 
    \end{subfigure}
    \begin{subfigure}{0.22\textwidth}%
    \vspace{-3pt}
    \centering\captionsetup{width=.95\textwidth}%
        \includegraphics[width=0.95\textwidth]{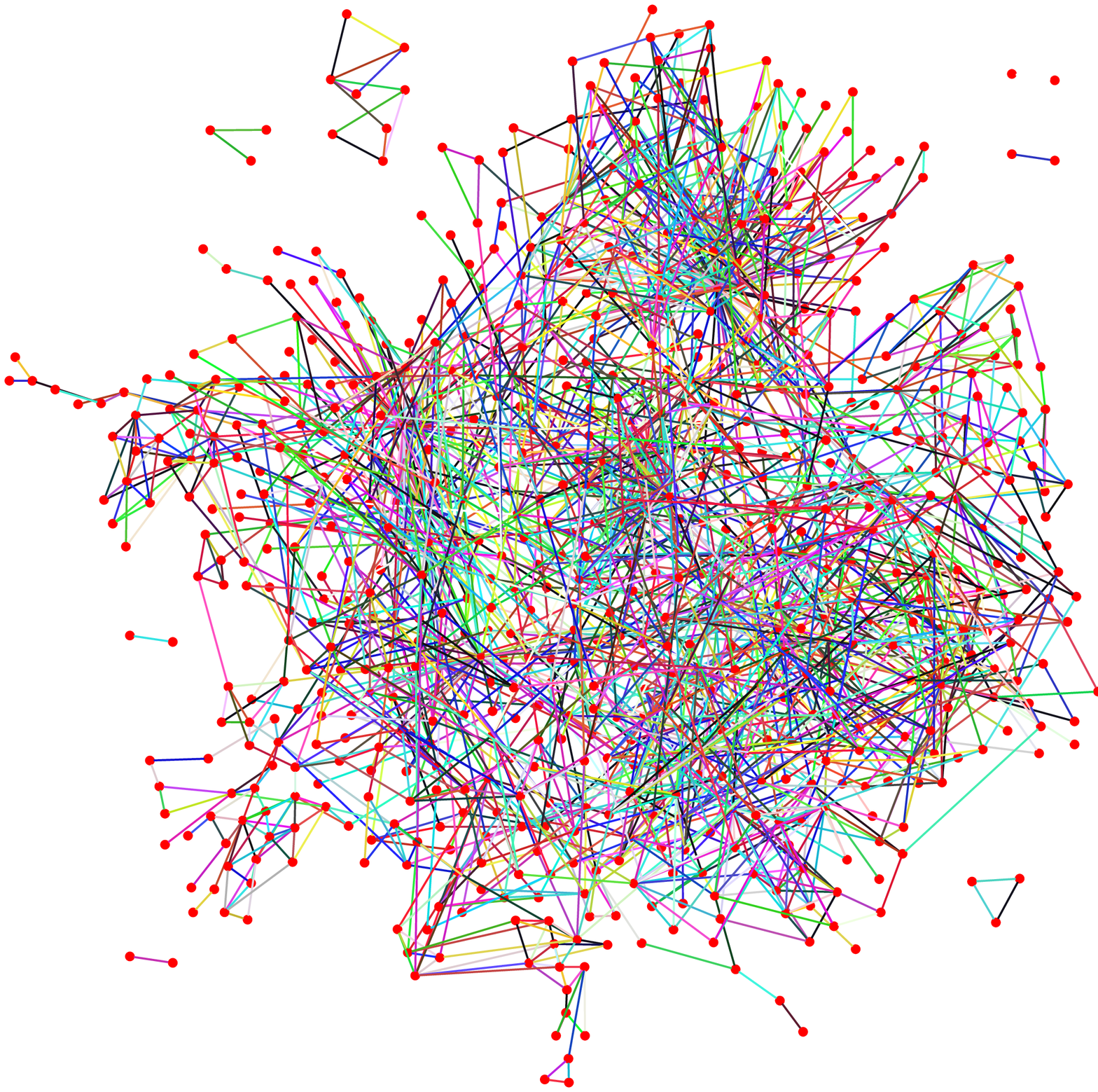}%
        \caption{Network synopsis.}
        \vspace{-4pt}
        \label{fig:cora:syn} 
    \end{subfigure}
    \caption{The synopsis of the network Cora with edge attribute information, represented in colors.}\vspace{-12pt}
    \label{fig:cora} 
\end{figure}

\textbf{Network with attribute information.} The H-GCN method exploits the deep learning method for network coarsening. The coarsening process executes structural equivalence grouping and collapsing followed by structural similarity grouping and collapsing. With the same set of neighbors, two nodes are considered to have structural equivalence. The structural similarity is computed based node and edge attribute information. After nodes get collapsed on structural equivalence, the collapsing based on structural similarity is carried out recursively until the whole network is visited. We illustrate one synopsis of the Cora network in Figure~\ref{fig:cora}. Cora is also a network with disconnected components. The original network is visualized in Figure~\ref{fig:cora:ori}, while the synopsis generated by H-GCN is in Figure~\ref{fig:cora:syn}. In comparison to the network without attribute information of Figure~\ref{fig:arxiv}, figures in Figure~\ref{fig:cora} are colored, representing the different values of edge attribute.

The above synopsis generation methods have not taken the node label generation into account for the synopsis graph. For networks both with and without attribute information, we adopt the following node label assignment scheme. In each hierarchy of synopsis generation, when two or more nodes with the same label are collapsed into a single node in the synopsis, the generated node has also the same label. If the labels of the collapsed nodes are not the same, a label is randomly picked from the original nodes' labels for the synopsis node. This randomness will be amended in the following two-phase tuning process of JITuNE.
\subsection{Similarity Measurement}
\label{sec:sim}

Similarity measurement module is designed to find the synopsis with the smallest size that can guarantee an effective transfer of tuning knowledge to the original network. On the one hand, the performance of hyperparameter tuning has a high correlation with the number of sampled configurations in the hyperparameter space. On the other hand, it is too costly to run an NE algorithm on a large number of hyperparameter configurations, if the input is a large-scale network. It is thus the best to apply hyperparameter tuning on a network as small as possible. But a synopsis in too small a size would not provide enough knowledge for the whole network. Similarity measurement module is to balance the trade-off in the process.

To measure the similarity between the synopsis and the original graph, we adopt the KL-divergence between distributions with regard to graph structures. To model the structure-related distribution of a graph, we let $p_{ij}$ be the probability of node $v_i$ and $v_j$ having an edge, where $v_i, v_j\in G$. Thus, with $W=\mathop{\Sigma}_{(i,j)\in E}w_{ij}$, where $w_{ij}\in X_E$, we have the following:
\begin{equation}
p_{ij}=\frac{w_{ij}}{W}
\end{equation}
To enable the computation of KL-divergence, we extend $G'$ into \textbf{$G'_x$}, where $V'_x=V$. In the extension of $G'$, every node $v_c$ formed by collapsing will be added with a mirror node $\overline{v}_c$, which has an edge connected to its original node $v_c$. This edge between $v_c$ and $\overline{v}_c$ has a unit weight. With $W'=\mathop{\Sigma}_{(i,j)\in E'}w'_{ij}$ for $G'$, where $w'_{ij}\in X_E'$, we can define a similar distribution for $G'_x$:
\begin{eqnarray}
q_{ij}=\frac{w'_{ij_x}}{W'_x}\\
W=W'+\Delta W\label{eq:ww'}\\
W'_x=W'+|V|-|V'|\label{eq:w'xw'}
\end{eqnarray}
Hence, the KL-divergence for the distributions defined on $G$ and $G'_x$ is represented as follows.
\begin{equation}
KL(\widetilde{G}||\widetilde{G}'_x)=\mathop{\Sigma}_{i,j}p_{ij}log\frac{p_{ij}}{q_{ij}}\label{eq:kld}\\
\end{equation}

The rationale behind extending $G'$ into $G'_x$ is as follows. Given the embedding $h_v'$ of a collapsed node in $G'$, we can obtain the embeddings $h_v$ of the corresponding original nodes in $G$ by adding a small variance vectors $u_\epsilon$ to $h_v'$, i.e., $h_v=h_v'+u_\epsilon$. In this way, we can get the embeddings of all the nodes in $G$ from the embeddings of $G'$. This is feasible because the synopsis generation process collapses \emph{similar} nodes. These nodes remain similar with embeddings generated in this form. And, if this extension generates useful embeddings, then the tuning on the synopsis graph is also applicable to the original graph. Henceforth, the feasibility condition for this extension is that the KL-divergence computed by Eq.\eqref{eq:kld} is not too large, i.e., the extended graph of $G'$ remains similar to the original graph $G$. Consider Eq.\eqref{eq:kld} again, we have:
\begin{equation}
KL(\widetilde{G}||\widetilde{G}'_x)=\mathop{\Sigma}_{i,j}p_{ij}\log \frac{w_{ij}(W-\Delta W+|V|-|V'|)}{w_{ij_x}W}\label{eq:size}\\
\end{equation}
Considering Eq.\eqref{eq:ww'} and Eq.\eqref{eq:w'xw'}, the free variables for KL-divergence in Eq.\eqref{eq:size} include $\Delta W$ and $|V'|$, both of which are related to $|V'|$, i.e., the size of the synopsis graph.

In a time-constrained process, we need a fast and efficient computation of graph similarity. Different similarity measurements can be used, e.g., those based on graph spectrum~\cite{spectral} or NetLSD~\cite{netlsd}, but such graph similarity computations tend to take a long time, violating the requirement of time constraint. To make JITuNE practical, we employ the network size ratio between the synopsis and the original graph as the similarity measurement, according to Eq.\eqref{eq:size}. In fact, the network size is a key metric to the performance of NE algorithms and affects the choice of hyperparameter values~\cite{deepwalk,node2vec}.

JITuNE generates the series of hierarchical network synopses and computes their similarities with the original graph. For the series of generated hierarchical synopses, we choose one at a certain level based on its similarity with the original graph. If the similarity is above a threshold, the synopsis for the next level is generated. This process is continued until the threshold is violated. For extremely large networks, the threshold must be computed progressively based on the time constraint and the synopsis generation times. Though simple and easy, our extensive experimental results in Section~\ref{sec:exp} validate our choice of similarity metric.
\subsection{Hyperparameter Tuning}
\label{sec:tune}

The time-constrained hyperparameter tuning process of JITuNE can be applied to both the network synopsis and the original network. The pseudocode for this flexible tuning process is listed in Algorithm~\ref{algo:jitune}. The tuning process is divided into two phases. The first phase is to tune the hyperparameter for the network synopsis and find a promising subspace of hyperparameters, while the second is to tune the hyperparameter on the original network $G$ within the found subspace. According to Definition~\ref{def:perf}, the result of hyperparameter tuning is related to the algorithm $M$, the network $G$ and the application $A$. These are the input of the hyperparameter tuning process.

At the beginning, the NE algorithm under tune is run on the original network once for acquiring the running time $t_G$. Given a duration $T$, we compute the rounds $r$ that can be used to run $M$ on $G'$ for collecting tuning samples. Let $R=\llcorner T/t_G\lrcorner$, the round number $r$ is deduced based on the following equations:\vspace{-3pt}
\begin{equation}
T=(1+\llcorner R/2\lrcorner)t_G+r\times t_G\times \rho\label{eq:time}\vspace{-3pt}
\end{equation}
Here, we approximate the algorithm runtime on $G'$ by $t_G$ with a ratio of $\rho$, where $\rho$ is a ratio concerning the time complexity of algorithm $M$, e.g., $\rho=\frac{|V'|\log|V'|}{|V|\log|V|}$~\cite{deepwalk}, $\rho=\frac{|E'|+|V'|}{|E|+|V|}$~\cite{arope}, or $\rho=\frac{|E'|}{|E|}$~\cite{gcn,line}.
\setlength{\textfloatsep}{0.6\baselineskip}
\begin{algorithm}[!t]
\footnotesize
\caption{The time-constrained hyperparameter tuning process for JITuNE}%
\label{algo:jitune}%
\KwIn{Network $G$; synopsis $G'$; network embedding algorithm $M$; time constraint $T$; the running time of $M$ on $G$ $t_G$.}
\KwOut{The optimal hyperparameter configuration $\theta_{opt}$.}
\tcc{Phase 1}
Compute round number $r$ based on Eq.~\eqref{eq:time}\;
Sample $r$ hyperparameter configurations into set $S_{\theta1}$\;
Run M on $G'$ with $S_{\theta1}$ and select the best $\theta'$\;
Trim hyperparameter space around $\theta'$\;
Sample $\llcorner R/2\lrcorner$ hyperparameter configurations into set $S_{\theta2}$\;
Run M on $G$ with $S_{\theta2}$ and select the best $\theta$\;
$\theta_{opt}\leftarrow \theta$\;
\tcc{Phase 2}
$T_{extra}\leftarrow checkExtraTime()$\;
\While{$T_{extra}>0$}{
    $r_{extra}=T_{extra}/t_G$\;
    Trim hyperparameter space around $\theta_{opt}$\;
    Sample $r_{extra}$ hyperparameter configurations $S_{\theta_e}$\;
    Run M on $G$ with $S_{\theta_e}$ and select the best $\theta$\;
    $\theta_{opt}\leftarrow \theta$\;
    $T_{extra}\leftarrow checkExtraTime()$\;
}
\Return $\theta_{opt}$\;
\end{algorithm}

The core process of hyperparameter tuning exploits a sample-and-search method~\cite{optuna,autotune,bestconfig}, which can effectively constrain its time used for the optimization process. The method divides the tuning process into two parts. In the first part, $r$ samples are taken from the hyperparameter space. The sampling method must guarantee a good coverage of the hyperparameter space for a good tuning result. Among common sampling methods, the Latin hypercube sampling (LHS) method is favored~\cite{bestconfig} and has demonstrates superior performance in hyperparameter tuning with a limited sample set. Then the best configuration $\theta'$ is selected from the $r$ samples, with the hyperparameter space trimmed around $\theta'$ using methods from \cite{optuna,bestconfig}. For a synopsis $G'$ with a similarity $\alpha$ to $G$, we trim each hyperparameter range to $1-\alpha$ of its original range to generate the hyperparameter subspace. Note that, \emph{there exists a trade-off here in choosing the similarity metric $\alpha$}. A small $\alpha$ can increase the sample number for hyperparameter tuning on the synopsis graph, but it will also increase the volume of the generated subspace for tuning on the original graph. Finally, $\llcorner R/2\lrcorner$ samples are taken in the hyperparameter subspace and evaluated on the original graph of $G$. The best performing configuration $\theta_{opt}$ is selected and output as the final optimal solution.

JITuNE takes two main measures to constrain the time of hyperparameter tuning. One is to reduce the size of input data, and the other is to decrease the samples in need. To achieve the first goal, JITuNE proposes to tune over the hierarchical synopsis of massive-scale networks. To achieve the second goal, JITuNE exploits a sampling method with high coverage over space. Besides, due to the impossibility of modeling on a few samples, JITuNE adopts the classic search-based method for hyperparameter optimization. On users given abundant time, the framework can be flexibly switched to learning-based methods that can transfer tuning knowledge from synopsis graphs to the original graph.

\section{Experimental Evaluation}%
\label{sec:exp}

In this section, we empirically evaluate our framework of JITuNE. Networks with and without attribute information are both considered. We demonstrate the efficacy and the efficiency of JITuNE on three categories of NE algorithms, i.e., sampling-based algorithms, factorization-based algorithms, and deep neural network-based algorithms. Regarding the applications, we optimize towards link prediction and node classification of networks.
\begin{table}[!b]
    \centering
    \footnotesize
    \vspace{6pt}
    \caption{Network datasets.}\vspace{-6pt}%
    \label{tbl:datasets}%
    \begin{tabular}{cccccc}
  \toprule[1.2pt]
    {Dataset} & {Category} &{\#Nodes}&{\#Edges}&{\#Labels}&{w/ attr.}\\
    \midrule[0.8pt]
    Blog.\cite{FlickrBlogcatalog}&social&10,312&333,983&39&F\\
    Wiki.\cite{wiki}&word&4,777&184,812&40&F\\
    Pubmed\cite{Pubmed}&citation&19,717&44,338&3&T\\
    \midrule[0.2pt]
    Arxiv\cite{arxiv}&collab.&5,242&28,980&--&F\\
    Cora\cite{cora}&citation&2,708&5,429&7&T\\
    \midrule[0.2pt]
    Flickr\cite{FlickrBlogcatalog}&social&80,513&11,799,764 &195&F\\
    Topcat\cite{Topcats}&web&1,791,489&28,511,807&17,364&F\\
  \bottomrule[1.2pt]
  \end{tabular}
  \end{table}
\subsection{Baselines and Experiment Settings}

The baselines we tested include random search and Bayesian optimization. Random search is the most commonly used hyperparameter tuning method. It is shown that random search can achieve noticeable improvements over default hyperparameter configurations~\cite{rsopt}. Baysian optimization is also commonly used when the function to optimize is unknown~\cite{bogpr}. It exploits the fact that the space for the unknown function constitutes a Gaussian stochastic process. With an appropriate utility function, the optimal point can be computed on a limited number of samples. A state-of-the-art work AutoNE~\cite{autone} is also compared. It is the work closest to ours. One of the major contribution of AutoNE is a component named meta-learner for transferring tuning knowledge from sampled sub-networks to the original network. We find that in quite a few cases, AutoNE performs much better \emph{without} the meta-learner component than with. Therefore, we also take AutoNE without meta-learner as a comparison choice.

We evaluate on the same set of network embedding algorithms as our counterparts, i.e., AROPE~\cite{arope}, DeepWalk~\cite{deepwalk} and GCN~\cite{gcn}. They are chosen to represent three typical categories  of NE algorithms. The hyperparameters for tuning are those having real impacts on performances and chosen according to the respective papers. As for tuning AROPE, we tune the weights of its first-order($w1$), second-order($w2$), and third-order($w3$) proximities, setting their ranges to $[1e-4, 3]$. For DeepWalk, the tuned hyperparameters include the number of random walks ($[40, 100]$), the length of each random walk ($[20, 80]$), the window size ($[5, 30]$), and the embedding size ($[40, 256]$). The hyperparameters tuned for GCN are the number of training epochs ($[10, 300]$), the number of neurons for each hidden layer ($[2, 64]$), the learning rate ($[1e-4, 0.1]$), the dropout rate ($[0.1, 0.9]$), and the weight decay for L2 regularization ($[1e-5, 1e-3]$). To demonstrate JITuNE's capability in tuning categorical hyperparameters along with numerical hyperparameters, we also tune the model types of GCN, i.e., choosing models from the set \{'gcn', 'gcn\_cheby', 'dense'\}.
\begin{figure*}[t]
    \begin{subfigure}{0.48\textwidth}%
    \centering%
        \includegraphics[width=.95\textwidth]{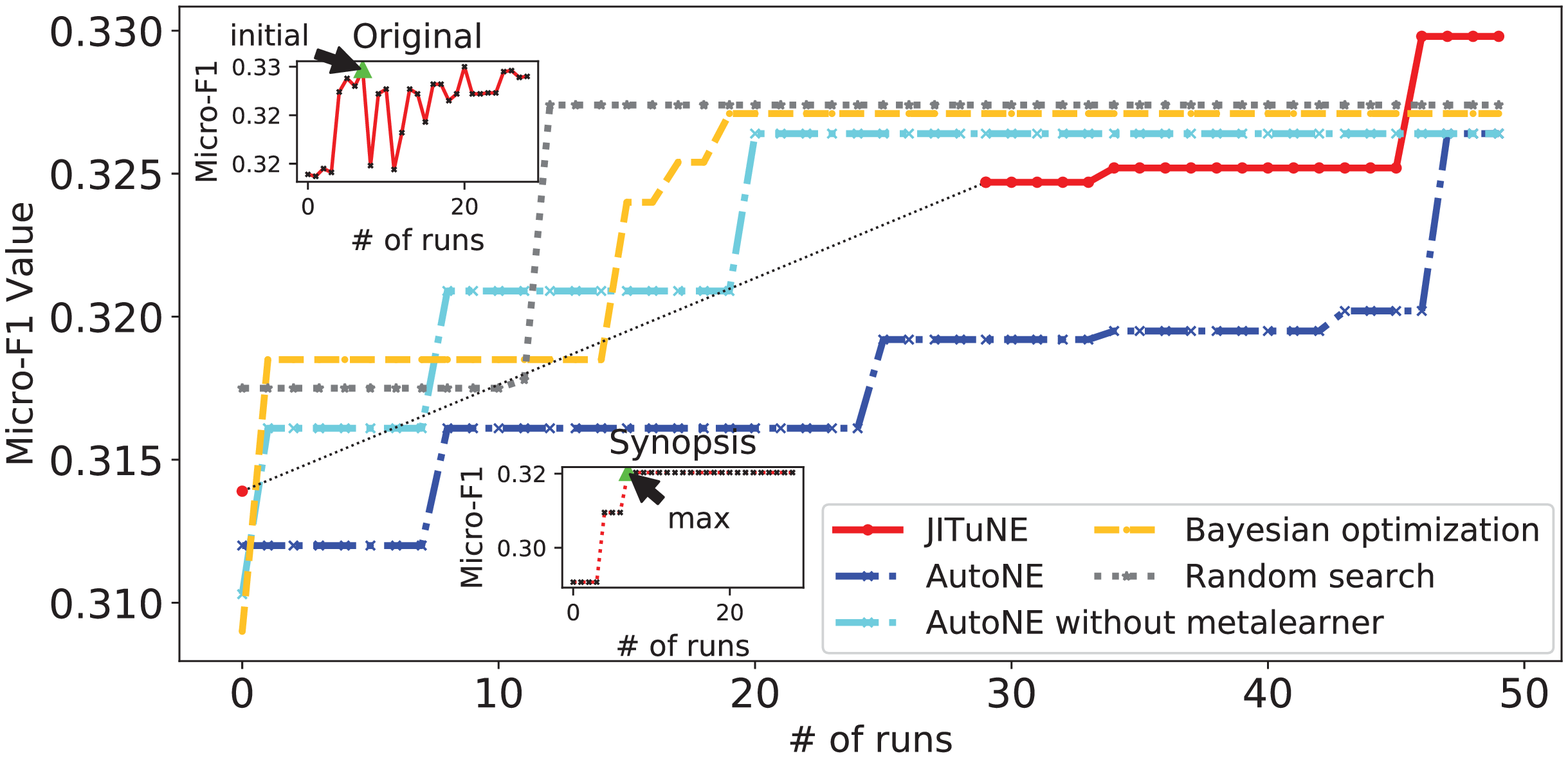}\vspace{-3pt}%
        \caption{Classification: Blog}\vspace{-3pt}
        \label{fig:arope:cb} 
    \end{subfigure}
    \begin{subfigure}{0.48\textwidth}%
    \centering%
        \includegraphics[width=.95\textwidth]{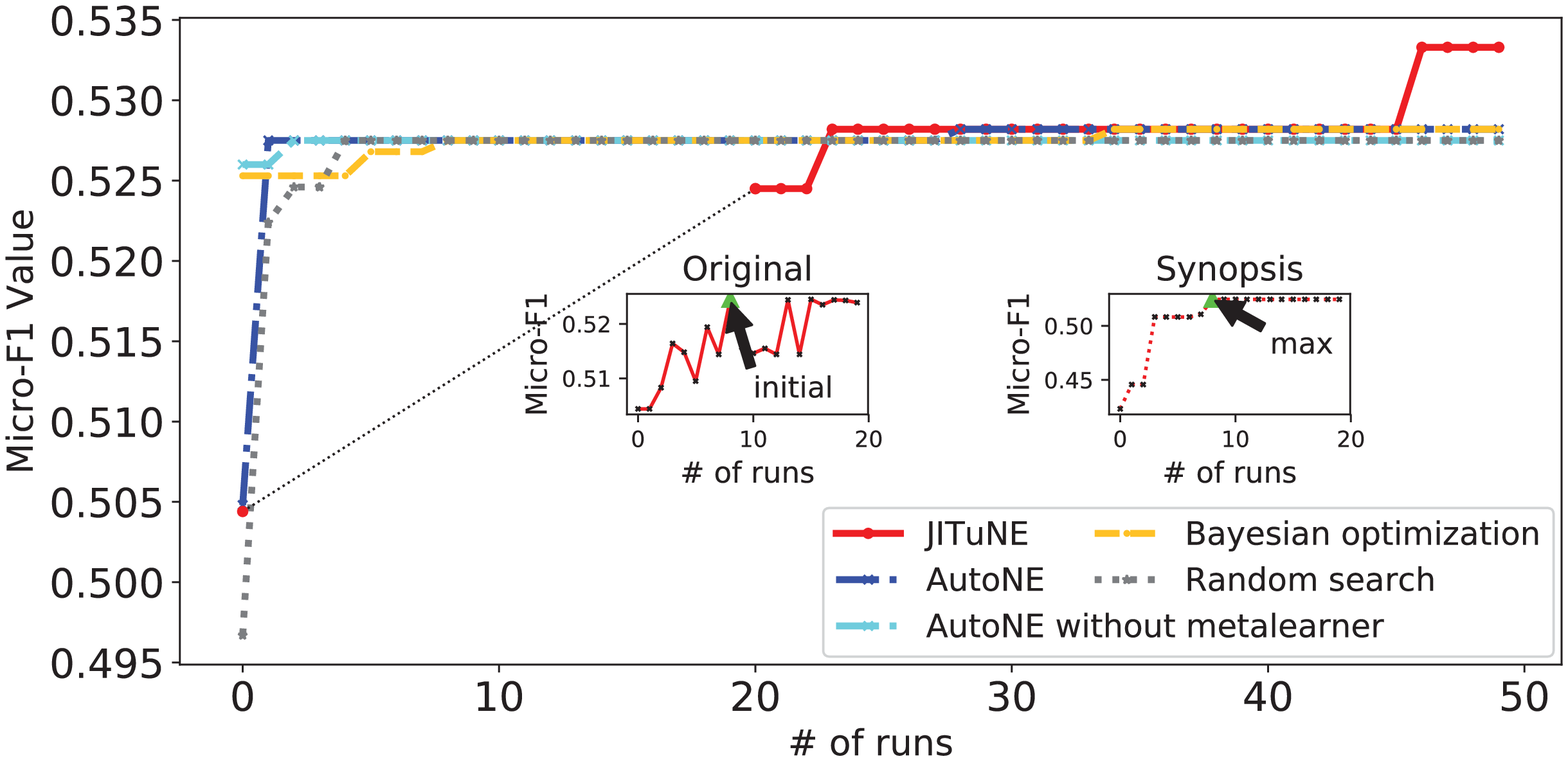}\vspace{-3pt}%
        \caption{Classification: Wiki}\vspace{-3pt}
        \label{fig:arope:cw} 
    \end{subfigure}
    \begin{subfigure}{0.48\textwidth}%
    \centering%
        \includegraphics[width=.95\textwidth]{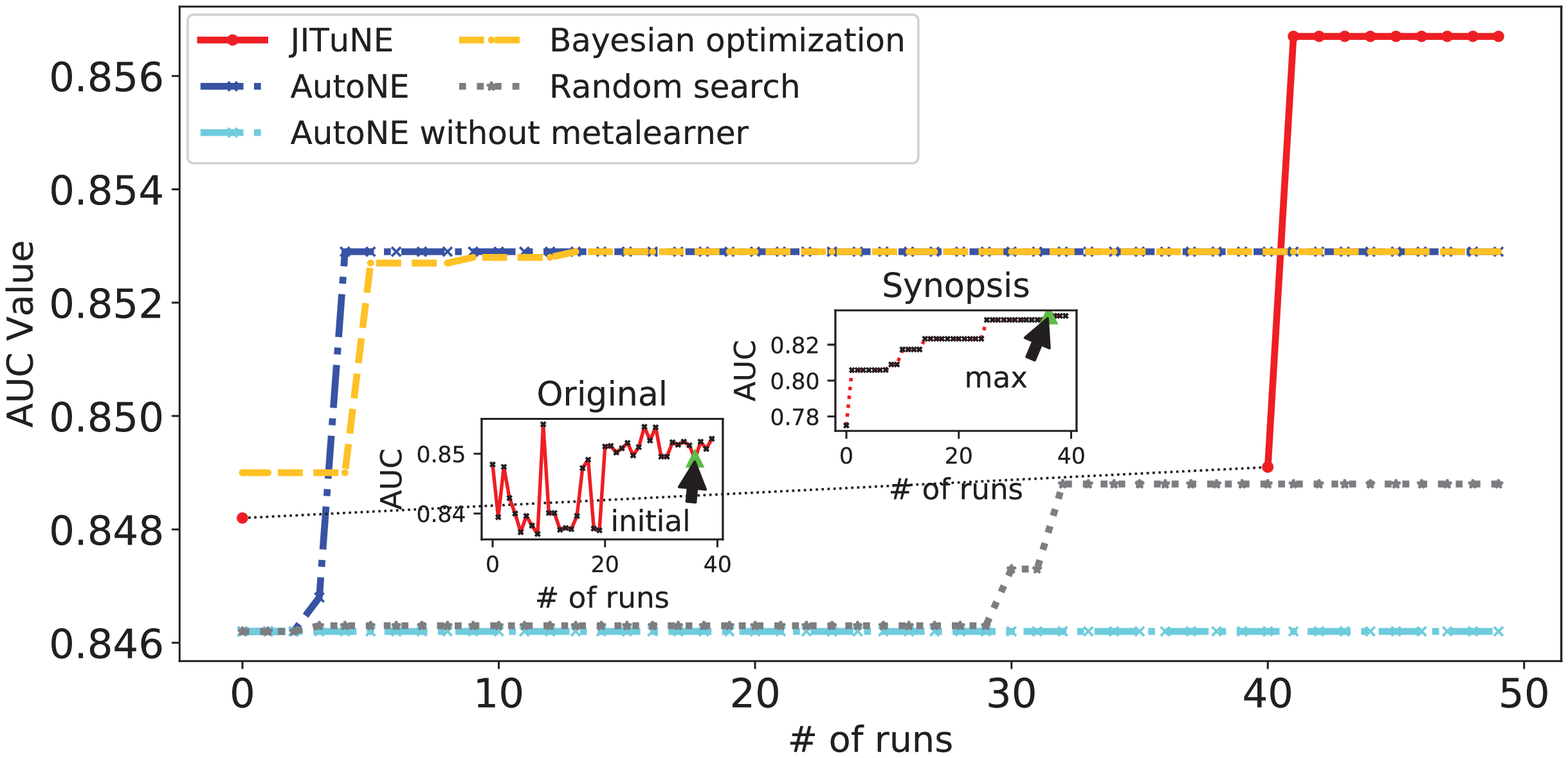}\vspace{-3pt}%
        \caption{Link prediction: Blog}\vspace{-3pt}
        \label{fig:arope:lb} 
    \end{subfigure}
    \begin{subfigure}{0.48\textwidth}%
    \centering%
        \includegraphics[width=.95\textwidth]{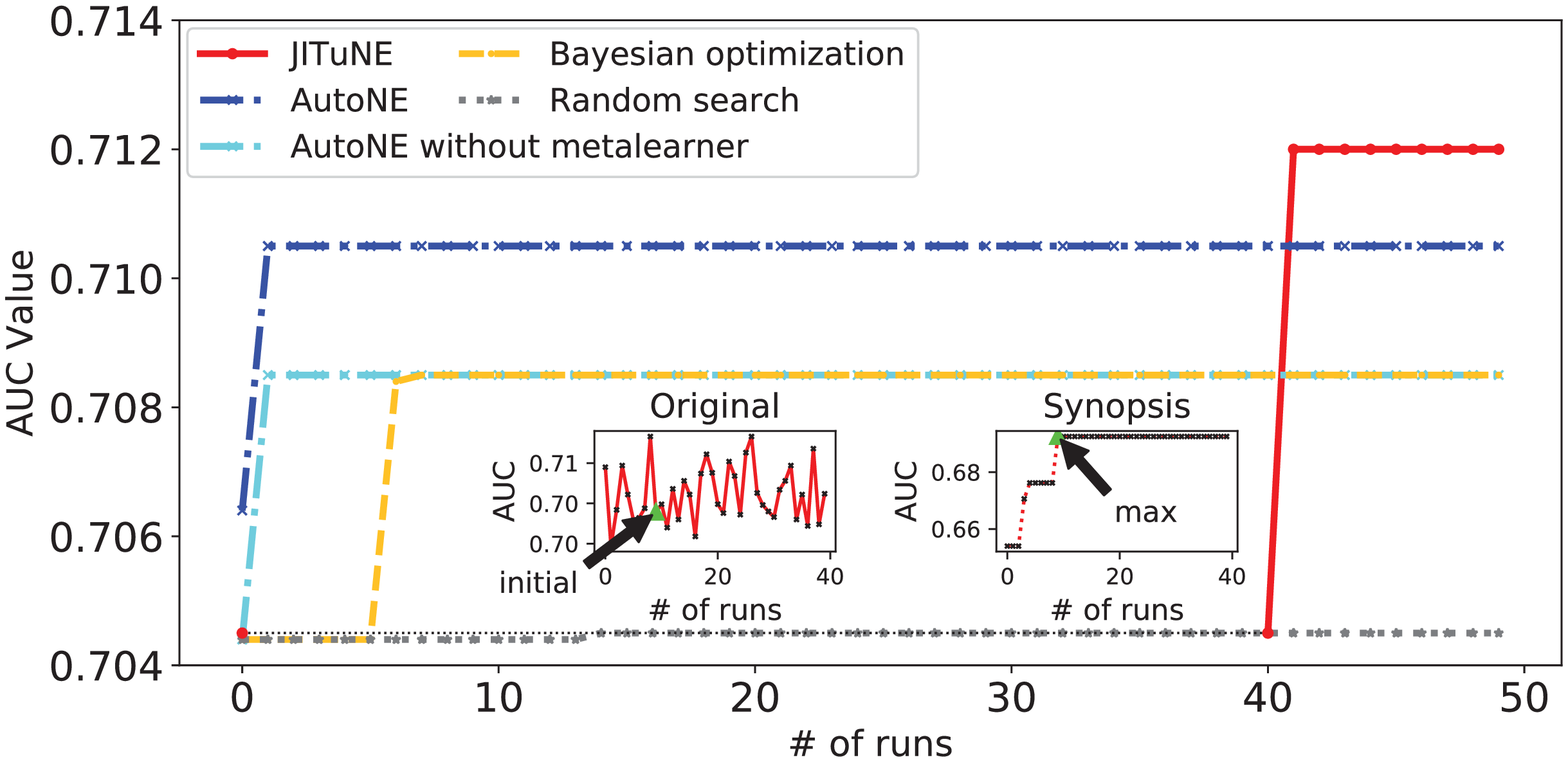}\vspace{-3pt}%
        \caption{Link prediction: Wiki}\vspace{-3pt}
        \label{fig:arope:lw} 
    \end{subfigure}
    \caption{Tuning results of different methods within the same time on the networks of BlogCatalog and Wikipedia. The NE algorithms under tune is AROPE. JITuNE spends part of the time on the synopsis graph, represented by a thinner dashed line. Two inset axes plot the performances for exploiting the corresponding hyperparameter configurations on the synopsis and on the original graph respectively.}\vspace{-6pt}
    \label{fig:arope} 
\end{figure*}
\begin{figure*}[t]
    \begin{subfigure}{0.48\textwidth}%
    \centering\captionsetup{width=.95\textwidth}%
        \includegraphics[width=\textwidth]{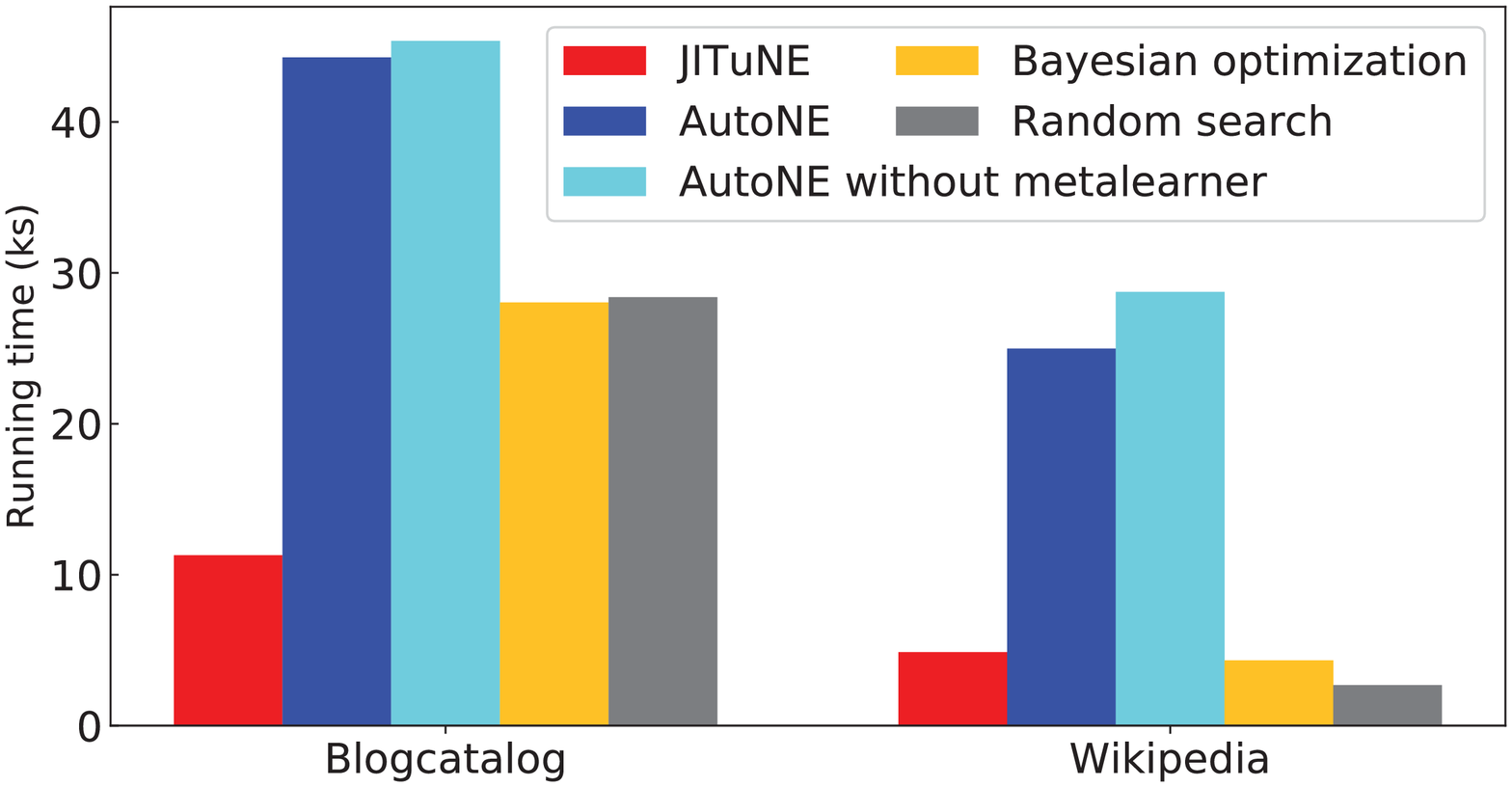}%
        \caption{Classification}\vspace{-3pt}
        \label{fig:aropeT:ct} 
    \end{subfigure}
    \begin{subfigure}{0.48\textwidth}%
    \centering\captionsetup{width=.95\textwidth}%
        \includegraphics[width=\textwidth]{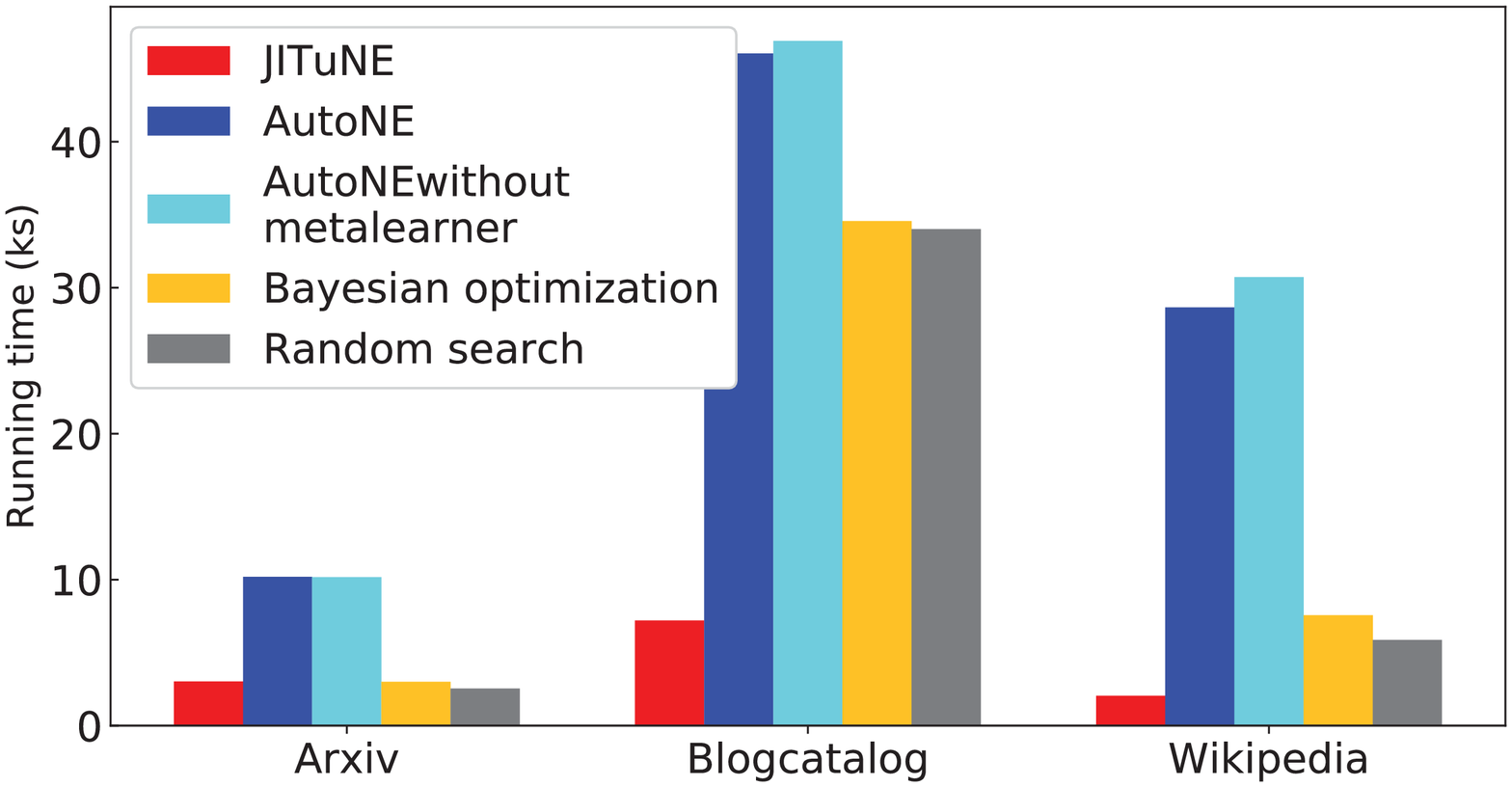}%
        \caption{Link prediction}\vspace{-3pt}
        \label{fig:aropeT:lt} 
    \end{subfigure}
    \caption{The total running time required by each hyperparameter optimization methods. The NE algorithms under tune are AROPE. The rounds for running AROPE are the same for all methods. The differences resulted in the final time are mainly due to the preprocessing and the computation time, e.g., the synopsis generation time for JITuNE and the time of sampling multiple sub-networks for AutoNE.}\vspace{-12pt}
    \label{fig:aropeT} 
\end{figure*}

Our experiments are carried out on datasets considered in a state-of-the-art related work~\cite{autone}. Although JITuNE can effectively tune the hyperparameters of NE algorithms on massive-scale networks, our counterparts cannot. Therefore, we first evaluate JITuNE against the baselines on small- and moderate-scale networks, wherein the baselines can finish in a reasonable time. To actually demonstrate the benefits of JITuNE, we validate our framework in two large-scale networks that no baselines can work as well in a reasonable time. For the ease of comparison, we select two large-scale networks from the related works~\cite{deepwalk,autone}. We summarize the information of the seven network datasets in our experiments in Table~\ref{tbl:datasets}. The datasets considers connected networks like BlogCatalog and disconnected networks like Arxiv.

We compare different tuning methods based on the final performances of the NE algorithms under tune. We measure the performance of each target NE algorithm with regard to two common applications, i.e., link prediction and node classification. In link prediction, we hold out $20\%$ of the edge set for test. The performance measurement for link prediction is the area under the curve (AUC)~\cite{auc}. In node classification, we train a logistic regression classifier using $20\%$ of the learned node representations as the input. The Micro-F1 score is used as the metric for evaluating AROPE and DeepWalk, while GCN is evaluated by classification accuracy.

To bound the tuning time, we let all the tuning methods run the same application under tune for 50 rounds. We can use algorithm rounds instead of actual running time because we are comparing these tuning methods with regard to the same target algorithm and application. Besides, Bayesian optimization takes five extra algorithm runs for initiating the model. For AutoNE and AutoNE without meta-learner, five extra algorithms runs are needed for tuning each of the five sampled sub-networks. Besides, time is needed for sampling sub-networks and computing transferred knowledge. In comparison, JITuNE counts the time for tuning synopsis and that for tuning the whole network in the 50 algorithm runs. We adopt a similarity threshold of $0.5$ for a better trade-off result according to discussion in Section~\ref{sec:sim} and Section~\ref{sec:tune}.

Our algorithms are implemented in Python 3.6, running on the operating system of Centos 7. All experiments are conducted on a machine with 192 GB memory and an Intel(R) Xeon(R) CPU E5-2630 v4 (2.2 GHz). The machine is equipped with GPU GeForce GTX 1080 Ti.\vspace{-12pt}

\subsection{Network w/o Attribute Information}

We first experiment with networks without attributes. We tune two types of NE algorithms, i.e., factorization-based and sampling-based algorithms. For each type, we test on both connected networks, e.g., BlogCatalog and Wikipedia, and a network with disconnected components, e.g., Arxiv. BlogCatalog is a social network, and Wikipedia is a co-occurrence network of words. Arxiv is a reference network. The network details are given in Table~\ref{tbl:datasets}.

\textbf{Factorization-based algorithms.} Methods in this category generally decompose the proximity matrix of graphs to acquire the network embedding representation. We tune the hyperparameters for the AROPE algorithm. We report the performance achieved by each method within the same number of rounds in Figure~\ref{fig:arope}. Some methods have extra computation steps, which lead to negligible time for tuning. We plot the total time required for each tuning method in Figure~\ref{fig:aropeT}.

Our framework significantly improves upon the baselines, as well as AutoNE. On the one hand, the hyperparameter configuration tuned for synopsis is already better than those optimized by other methods in most cases. On the other hand, with an extra validation and tuning phase over the original network, JITuNE can achieve even a better performance.

Notice that the performance results on the synopses are also plotted inside the performance figures as inset axes for references. Although the hyperparameters tuned for synopses do not work the best in the original network, they are good enough as the starting point for tuning the complete network. Besides, the hyperparameter tuning for synopses has pinned down effective range bounds for the tuning on the original network.
\begin{figure*}[t]
    \begin{subfigure}{0.48\textwidth}%
    \centering%
        \includegraphics[width=.95\textwidth]{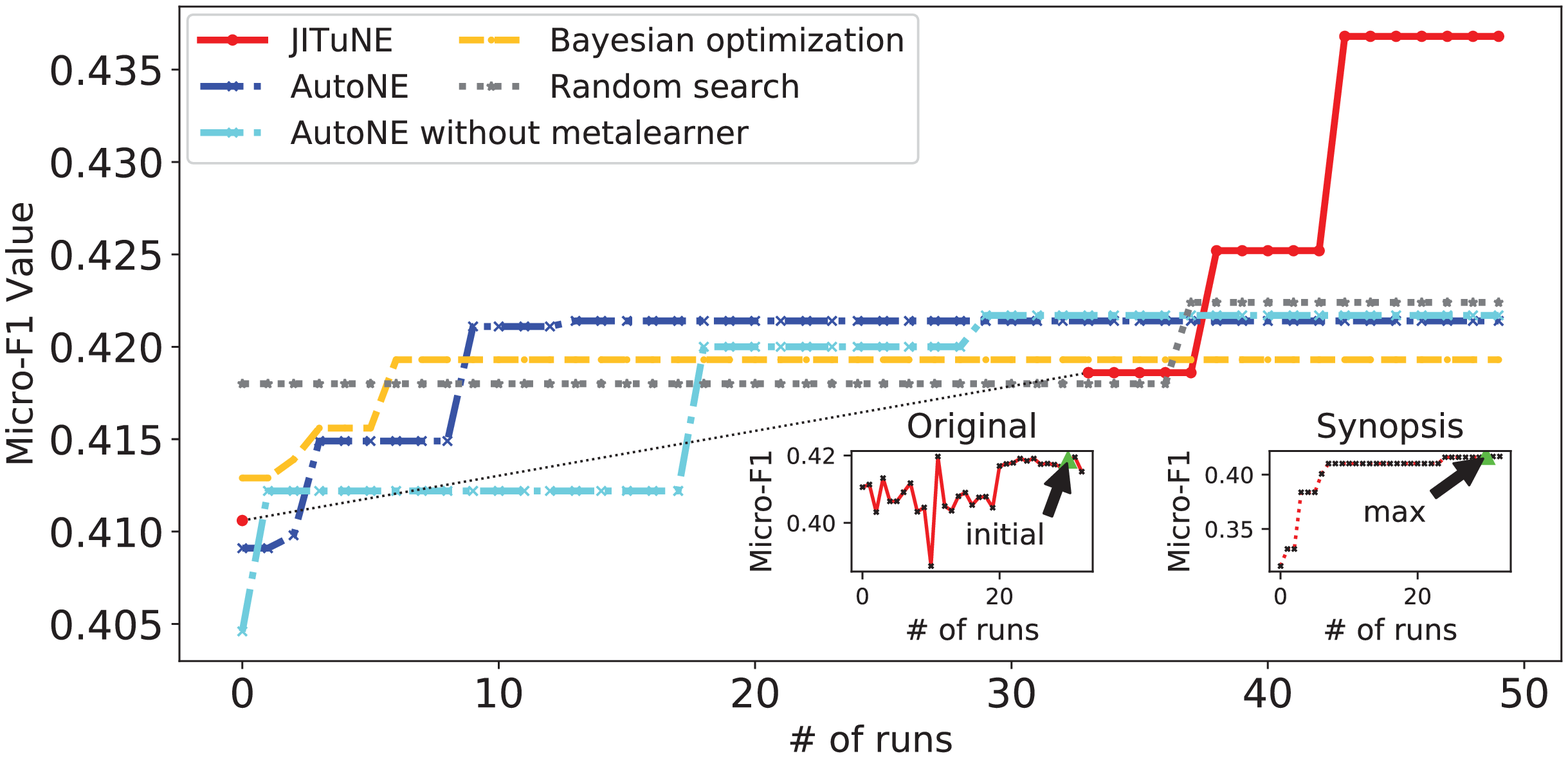}\vspace{-3pt}%
        \caption{Classification: Blog}\vspace{-3pt}
        \label{fig:deepwalk:cb} 
    \end{subfigure}
    \begin{subfigure}{0.48\textwidth}%
    \centering%
        \includegraphics[width=.95\textwidth]{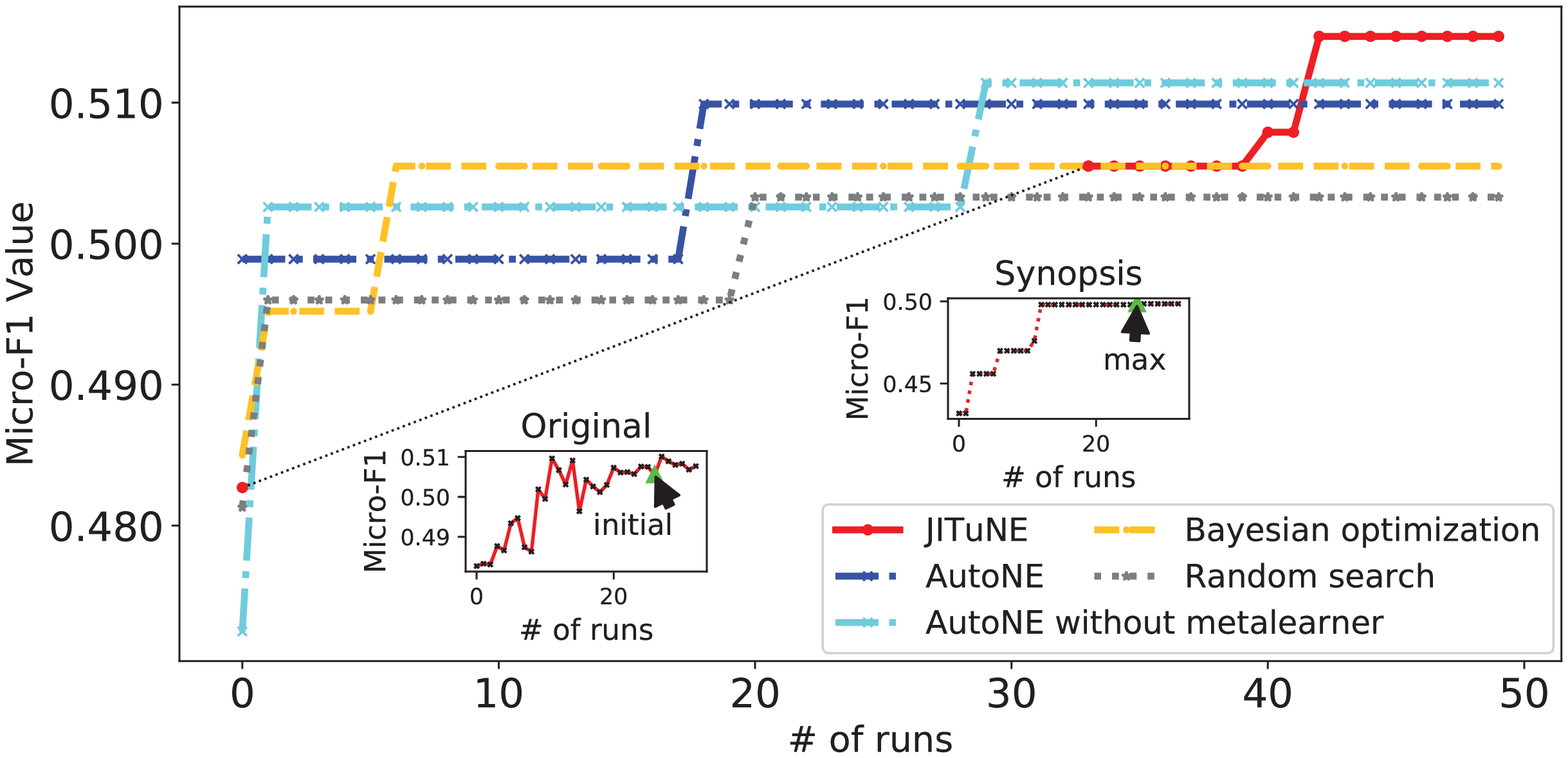}\vspace{-3pt}%
        \caption{Classification: Wiki}\vspace{-3pt}
        \label{fig:deepwalk:cw} 
    \end{subfigure}
    \begin{subfigure}{0.48\textwidth}%
    \centering%
        \includegraphics[width=.95\textwidth]{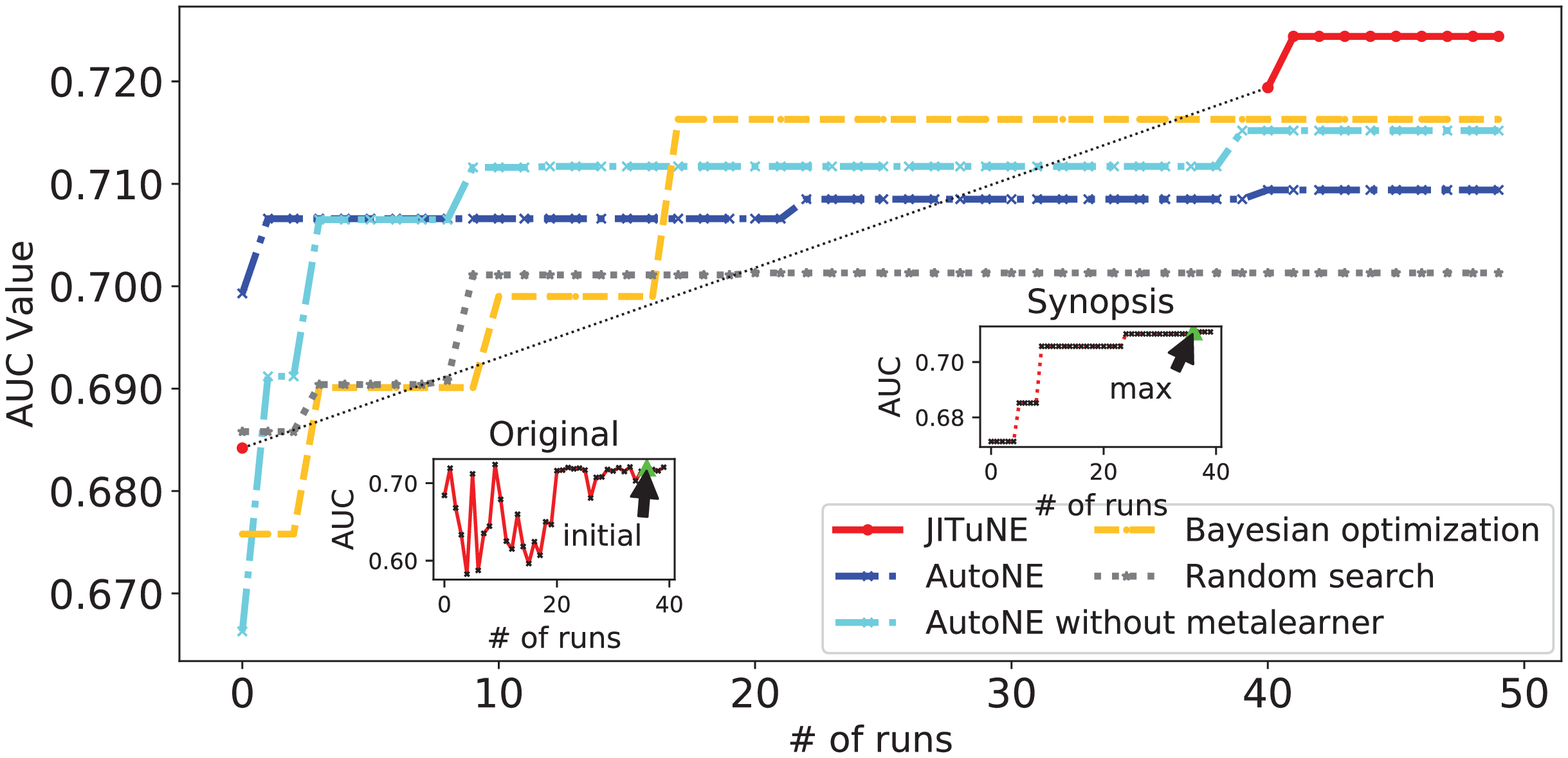}\vspace{-3pt}%
        \caption{Link prediction: Blog}\vspace{-3pt}
        \label{fig:deepwalk:lb} 
    \end{subfigure}
    \begin{subfigure}{0.48\textwidth}%
    \centering%
        \includegraphics[width=.95\textwidth]{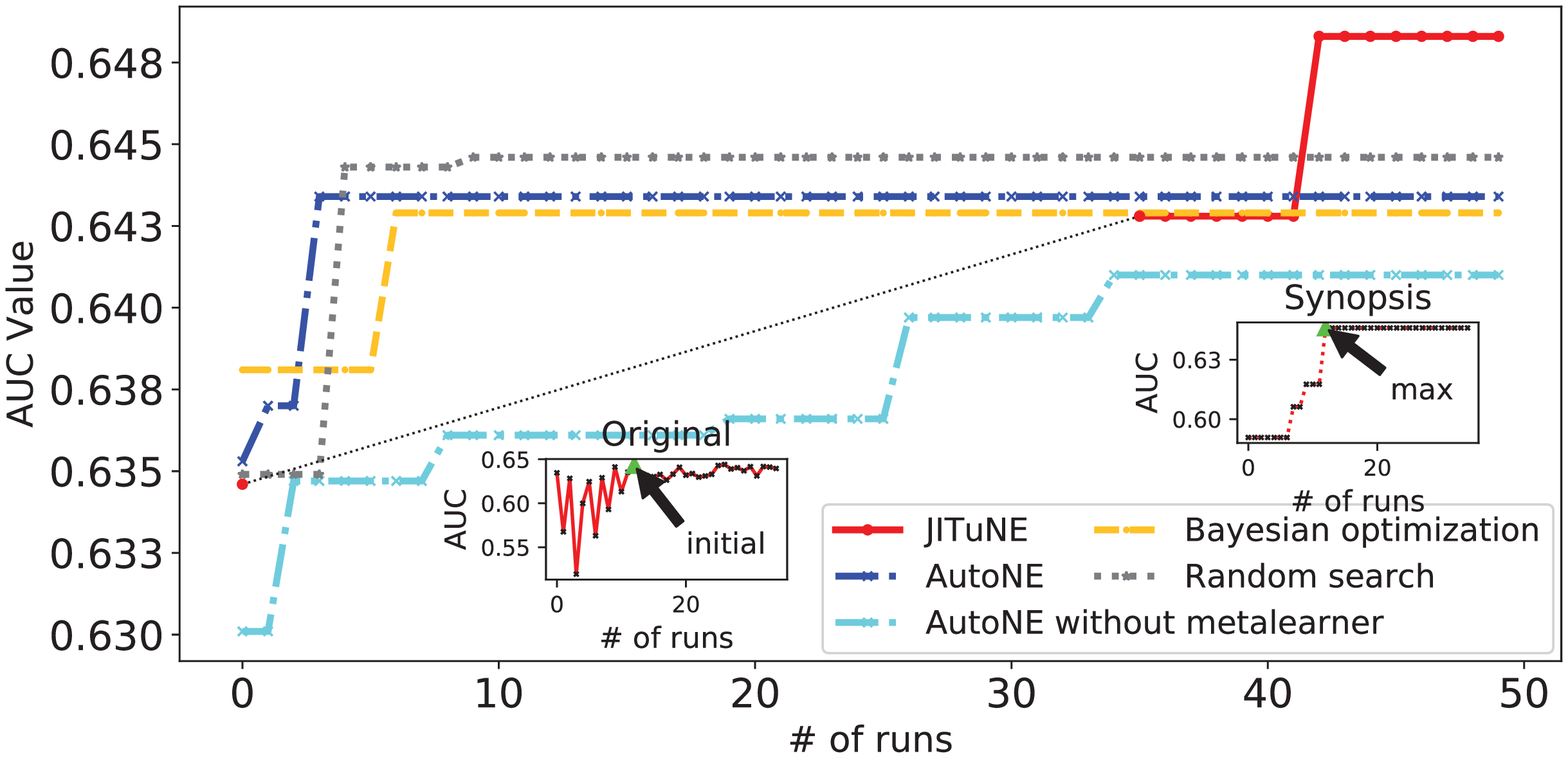}\vspace{-3pt}%
        \caption{Link prediction: Wiki}\vspace{-3pt}
        \label{fig:deepwalk:lw} 
    \end{subfigure}
    \caption{Tuning results of different methods within the same time on the networks of BlogCatalog and Wikipedia. The NE algorithms under tune is DeepWalk. JITuNE spends part of the time on the synopsis graph, represented by a thinner dashed line. Two inset axes plot the performances for exploiting the corresponding hyperparameter configurations on the synopsis and on the original graph respectively.}\vspace{-6pt}
    \label{fig:deepwalk} 
\end{figure*}
\begin{figure*}[t]
    \begin{subfigure}{0.48\textwidth}%
    \centering\captionsetup{width=.95\textwidth}%
        \includegraphics[width=\textwidth]{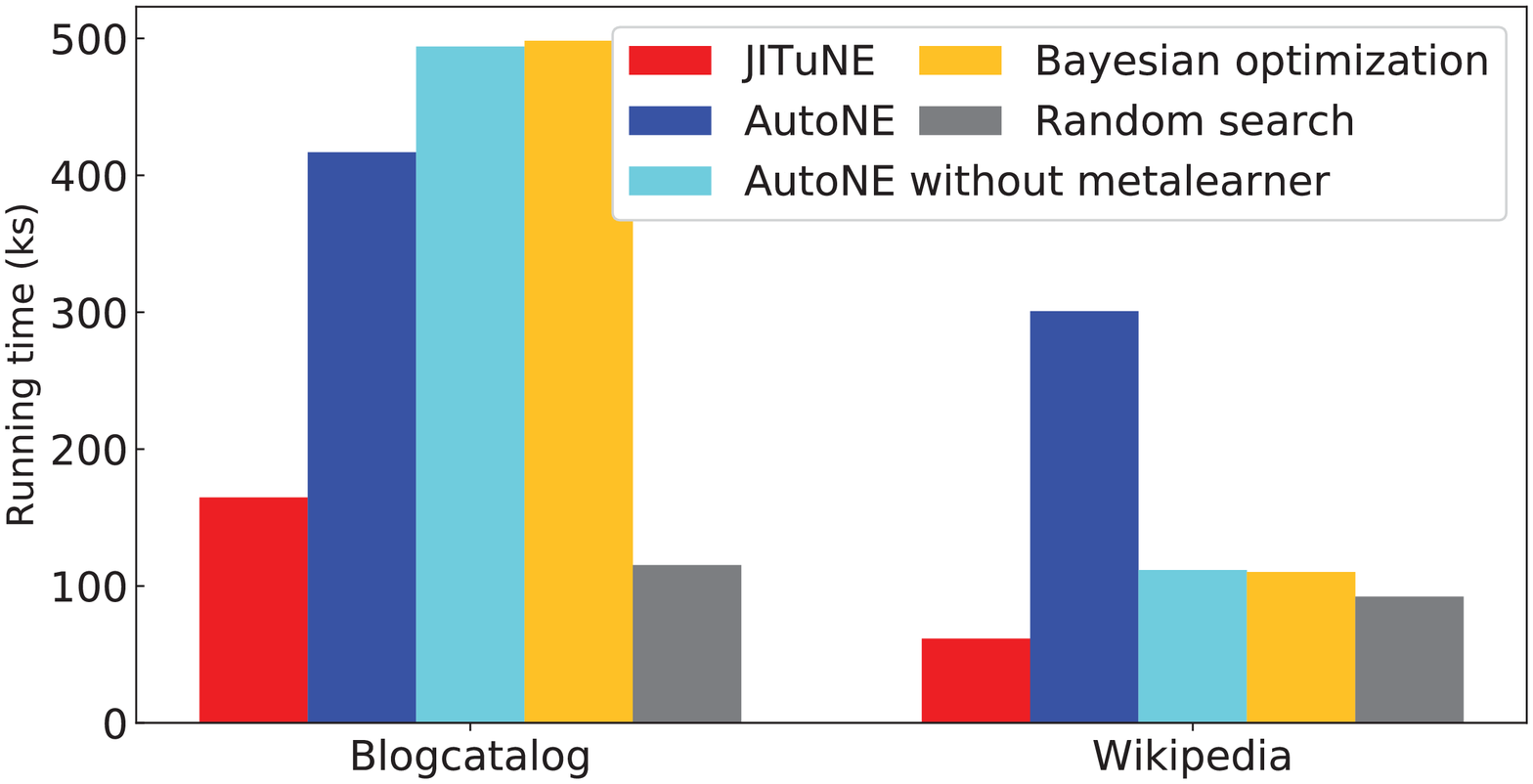}%
        \caption{Classification}\vspace{-3pt}
        \label{fig:deepwalkT:ct} 
    \end{subfigure}
    \begin{subfigure}{0.48\textwidth}%
    \centering\captionsetup{width=.95\textwidth}%
        \includegraphics[width=\textwidth]{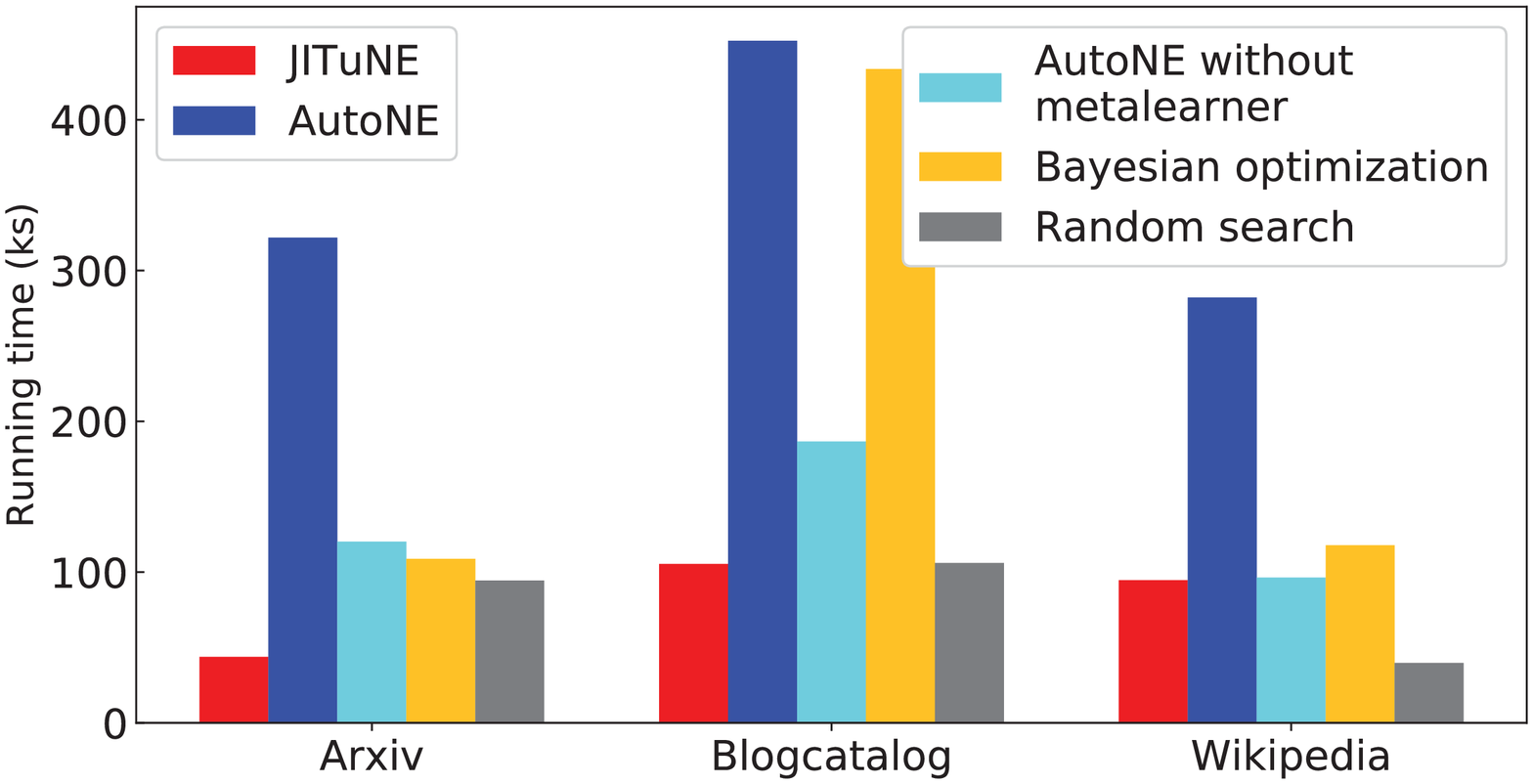}%
        \caption{Link prediction}\vspace{-3pt}
        \label{fig:deepwalkT:lt} 
    \end{subfigure}
    \caption{The total running time required by each hyperparameter tuning methods. The NE algorithms under tune are DeepWalk. The rounds for running DeepWalk are the same for all methods. The differences resulted in the final time are mainly due to the preprocessing and the computation time, e.g., the synopsis generation time for JITuNE and the time of sampling multiple sub-networks for AutoNE.}\vspace{-12pt}
    \label{fig:deepwalkT} 
\end{figure*}

\textbf{Sampling-based algorithms.} Methods in this category are inspired by the word2vec~\cite{word2vec} method of natural language processing. Random walk in the network is used to sample nodes and edges for learning node representations. Among the set of sampling-based algorithms~\cite{node2vec,line}, we test the DeepWalk algorithm~\cite{deepwalk}, which has four key hyperparameters, including the embedding size, the number of random walks starting  from each node, the length of each random walk and the window size.
\begin{figure*}[!t]
    \begin{subfigure}{0.48\textwidth}%
    \hspace{-6pt}
    \centering%
        \includegraphics[width=.95\textwidth]{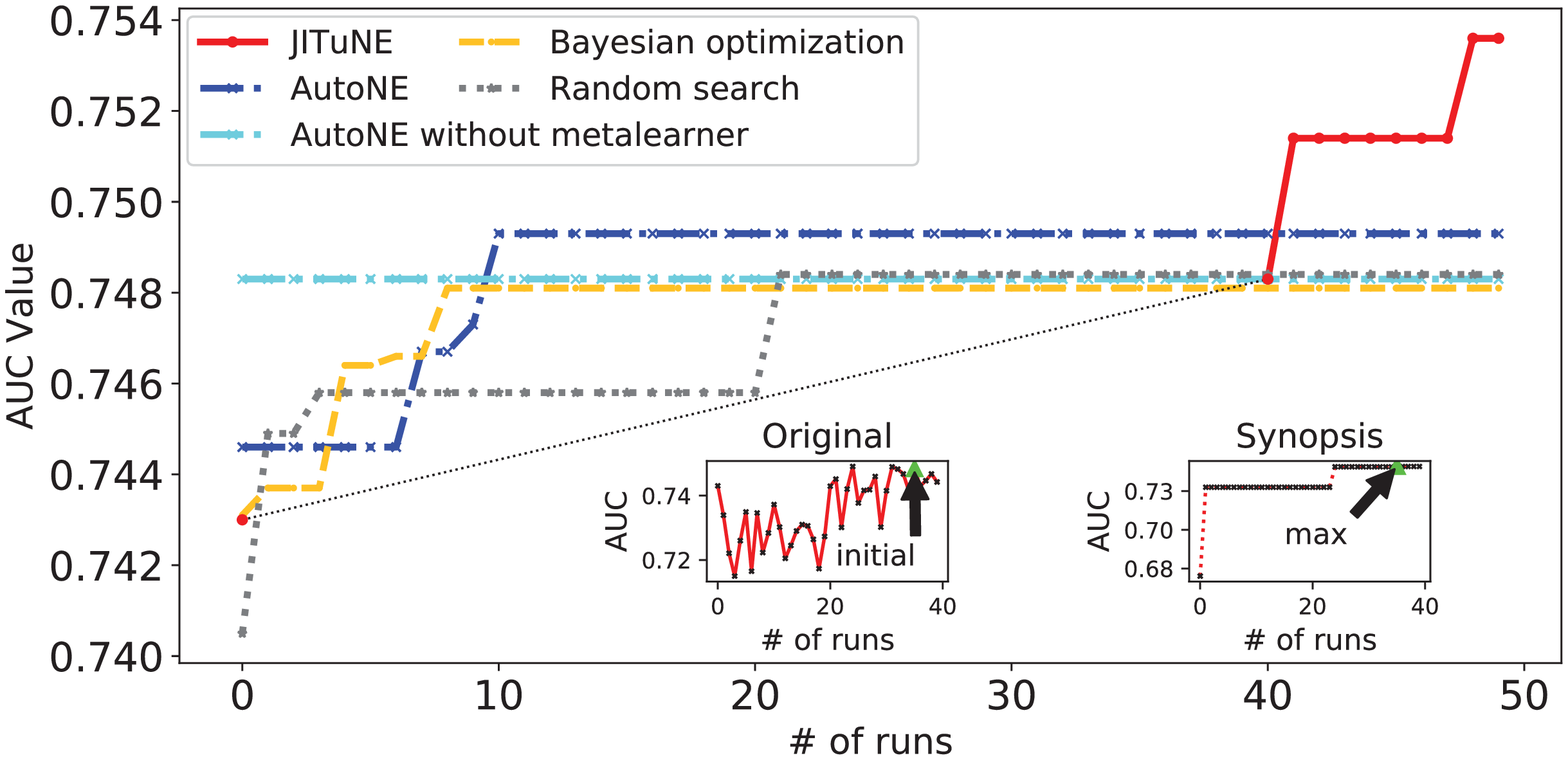}%
        \caption{AROPE}\vspace{-3pt}
        \label{fig:disnet:arope} 
    \end{subfigure}
    \begin{subfigure}{0.48\textwidth}%
    \centering%
        \includegraphics[width=.95\textwidth]{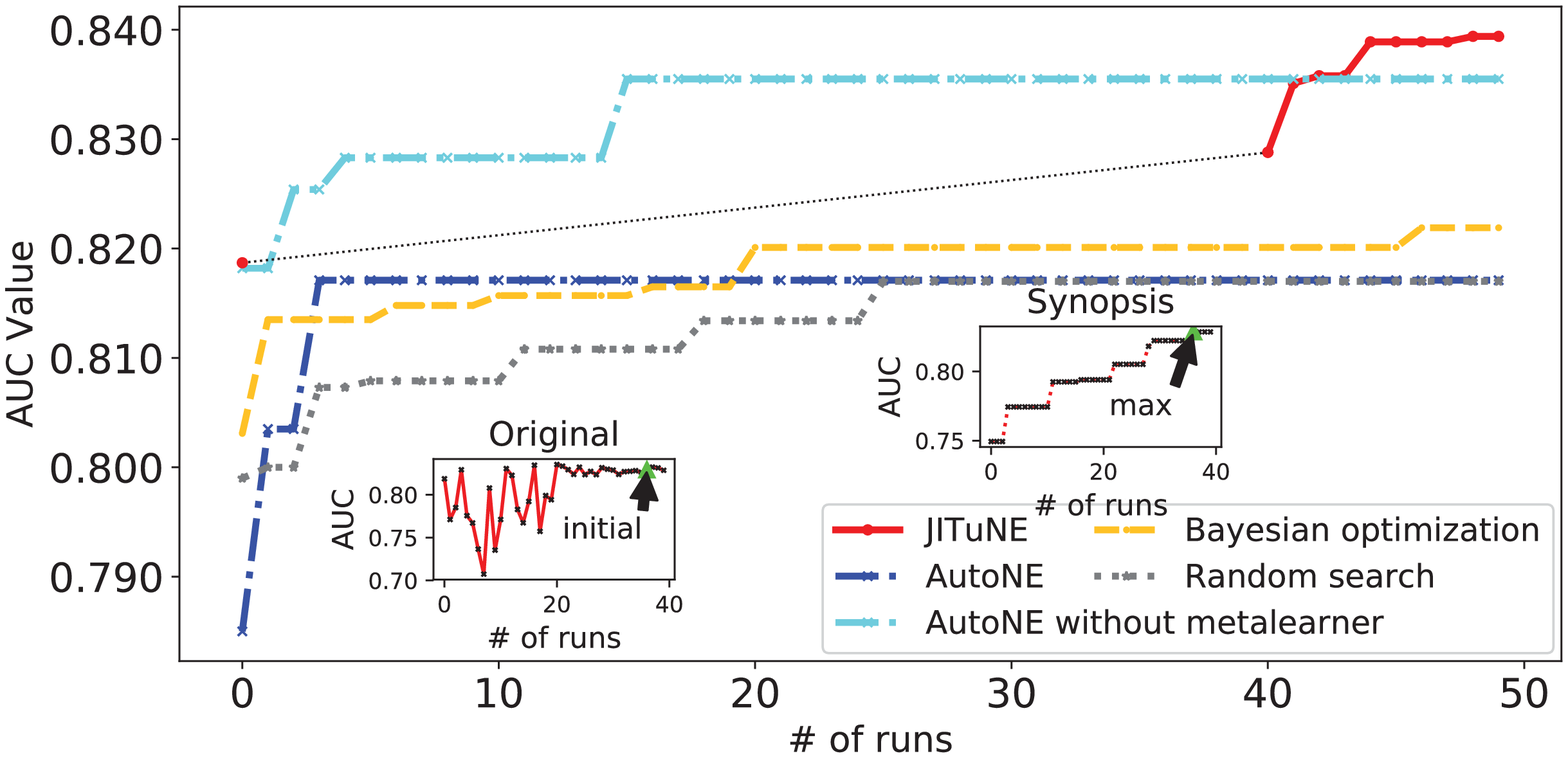}%
        \caption{Deepwalk}\vspace{-3pt}
        \label{fig:disnet:deepwalk} 
    \end{subfigure}
    \caption{Tuning results of different methods within the same time on the disconnected network of Arxiv. The NE algorithms under tune is AROPE and DeepWalk. The application is link prediction. JITuNE spends part of the time on the synopsis graph, represented by a thinner dashed line. Two inset axes plot the performances for exploiting the corresponding hyperparameter configurations on the synopsis and on the original graph respectively.}\vspace{-6pt}
    \label{fig:disnet} 
\end{figure*}
\begin{figure*}[t]
    \begin{subfigure}{0.48\textwidth}%
    \centering\captionsetup{width=.95\textwidth}%
        \includegraphics[width=\textwidth]{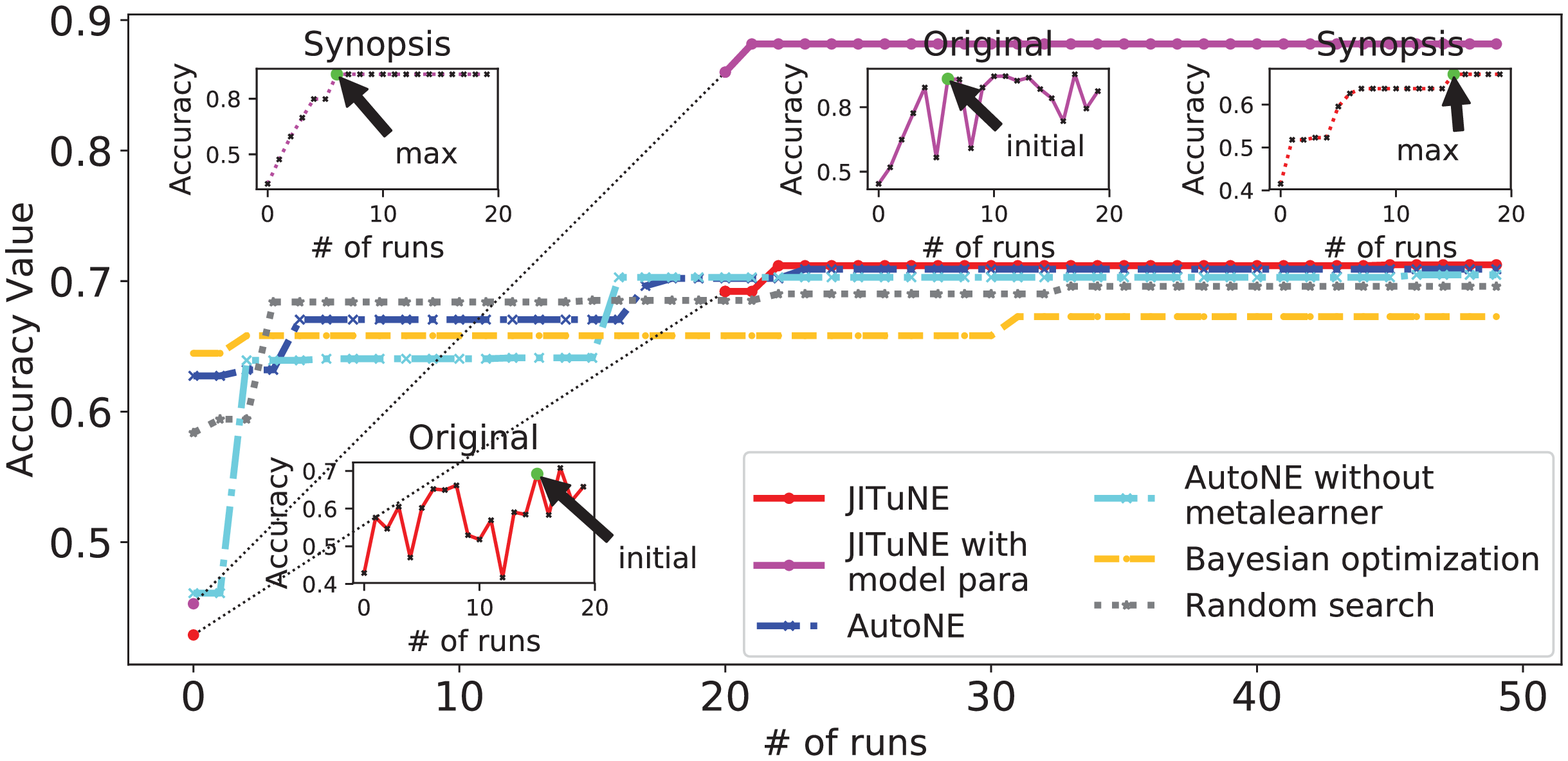}%
        \caption{Pubmed}\vspace{-3pt}
        \label{fig:gcn:pubmed} 
    \end{subfigure}
    \begin{subfigure}{0.48\textwidth}%
    \centering\captionsetup{width=.95\textwidth}%
        \includegraphics[width=\textwidth]{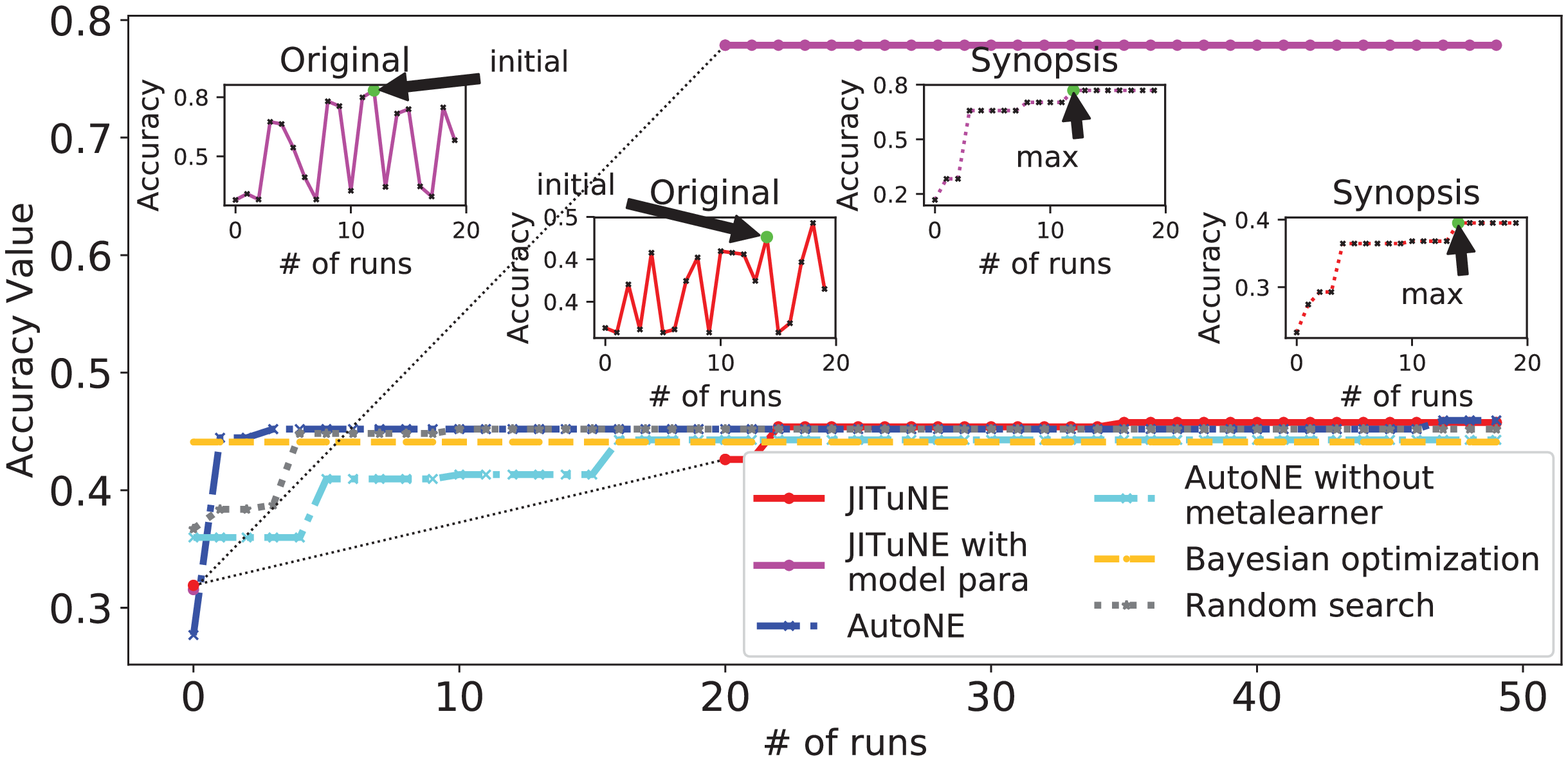}%
        \caption{Cora}\vspace{-3pt}
        \label{fig:gcn:cora} 
    \end{subfigure}
    \caption{Tuning results of different methods within the same time on the networks of Pubme and Cora. The NE algorithms under tune is GCN. JITuNE and JITuNE with model para spend part of the time on the synopsis graph, represented by thinner dashed lines. For either of the JITuNE cases, two inset axes plot the performances for exploiting the corresponding hyperparameter configurations on the synopsis and on the original graph respectively.}\vspace{-9pt}
    \label{fig:gcn} 
\end{figure*}

We measure the performance of hyperparameter optimization methods in terms of two metrics. The first is the performance tuned within a given number of rounds for running the NE algorithm. The second is the total time for running the hyperparameter optimization method. We report the results corresponding to these two metrics in Figure~\ref{fig:deepwalk} and Figure~\ref{fig:deepwalkT} respectively.

Figure~\ref{fig:deepwalk} and Figure~\ref{fig:deepwalkT} demonstrate that JITuNE can outperform the baselines and the recent work of AutoNE significantly and consistently, producing better results than the original NE algorithm papers~\cite{harp,deepwalk}. Besides, JITuNE spends little time in steps other than running NE algorithms. It has a running time much shorter than the recent work of AutoNE. Furthermore, JITuNE can take an even shorter time than random search and Bayesian optimization. This is accomplished through hyperparameter tuning on the synopsis, instead of on the original massive-scale network. In the sense of transfer learning, JITuNE is transferring knowledge from tuning the synopsis to tuning the raw data through a simple yet powerful method.

Besides the connected networks, the framework of JITuNE is also applicable to disconnected network. We experiment with the Arxiv dataset, which does not have label information. As a result, the classification application cannot be tested. We only tune DeepWalk and AROPE with regard to the link prediction application. The performance result is presented in Figure~\ref{fig:disnet}. JITuNE outperforms baselines and AutoNE to a large extent. Besides, as illustrated in Figure~\ref{fig:aropeT:lt} and Figure~\ref{fig:deepwalkT:lt}, the impressive performance is achieved in a shorter time than as required by other methods.\vspace{-6pt}

\subsection{Network w/ Attribute Information}

We also experiment with networks with attribute information. We tune the deep neural network-based algorithms on such networks. We test on a connected network, e.g., Pubmed, and a network with disconnected components, e.g., Cora. Pubmed and Cora are both citation networks. The network details are provided in Table~\ref{tbl:datasets}.

\begin{figure}[t]
    \centering
        \includegraphics[width=.45\textwidth]{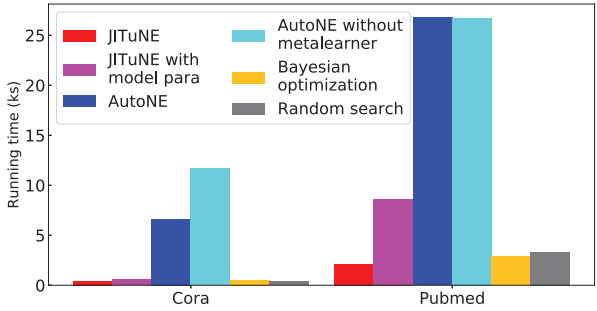}\vspace{-6pt}%
        \caption{The total running time required by each hyperparameter optimization methods. The NE algorithms under tune are GCN. The rounds for running GCN are the same for all methods. The differences resulted in the final time are mainly due to the preprocessing and the computation time, e.g., the synopsis generation time for JITuNE and the time of sampling multiple sub-networks for AutoNE.}
        \label{fig:gcnT} %
\end{figure}

\textbf{Deep neural network-based algorithms.} The fundamental idea of these methods~\cite{sdne,gcn,fastgcn} is to map the original network space to a low-dimensional vector space. The deep neural network-based method of GCN requires the input network have node/edge attribute information. We measure the performance of each method in node classification. We report the performance achieved by each method within the same number of rounds in Figure~\ref{fig:gcn}. On the one hand, JITuNE can achieve a performance no worse than any other method. On the other hand, GCN has a categorical hyperparameter that cannot fit into the framework of AutoNE but can be tuned by JITuNE. As illustrated in Figure~\ref{fig:gcn}, JITuNE can tune GCN to a much higher performance with this extra hyperparameter in consideration, as represented by the curve of \emph{JITuNE with model para}. What is even better is that JITuNE can tune GCN in a much shorter time than all other counterparts, as shown in Figure~\ref{fig:gcnT}, even when one more hyperparameter is added.\vspace{-6pt}

\subsection{Analysis on Large-Scale Networks}
\label{sec:exp:largeNet}

JITuNE can excel in hyperparameter tuning for NE algorithms within a time constraint. We now demonstrate the superiority of JITuNE in handling massive-scale networks. The details of the two large-scale datasets are  presented in Table~\ref{tbl:datasets}.
\begin{table}[!b]
    \centering
    \footnotesize
    \caption{Results on the large-scale network of Flickr, tuning algorithm DeepWalk for node classification.}\vspace{-3pt}%
    \label{tbl:flickr}%
    \begin{tabular}{ccccccc}
  \toprule[1.2pt]
  \multirow{2}{*}{\tiny Methods}&\multicolumn{2}{c}{\tiny Trail1}&\multicolumn{2}{c}{\tiny Trail2}&\multicolumn{2}{c}{\tiny Trail3}\\
\cline{2-7}
    &{\tiny Micro-F1} &{\tiny Time(s)}&{\tiny Micro-F1} &{\tiny Time(s)}&{\tiny Micro-F1} &{\tiny Time(s)}\\
    \midrule[0.8pt]
    JITuNE& \textbf{0.3028} & 2007.4 & \textbf{0.3132} & 3396.6 & \textbf{0.3341} & 4811.6\\
    AutoNE& 0.2839 & 3017.8 & 0.3018 & 4634.6 & 0.3297 & 6059.0\\
    Random& 0.2942 & 1438.8 & 0.2718 & 3057.8 & 0.3318 & 4574.6\\
  \bottomrule[1.2pt]
  \end{tabular}
  \end{table}

We first evaluate JITuNE on the large-scale network Flickr~\cite{FlickrBlogcatalog}, which is used by the DeepWalk paper~\cite{deepwalk}. We tune the DeepWalk algorithm and then test with the node classification application. The time constraint is set to three runs of DeepWalk algorithms on the original network, plus three runs on the network synopsis. The compared AutoNE framework execute three runs on five sampled sub-networks respectively before the three runs on the complete network. The random method involves only three runs on the original network. Table~\ref{tbl:flickr} presents the performance results along with the time cost, which sums up all the algorithm runs for each method.
\begin{table}[!b]
    \centering
    \footnotesize
    \caption{Results on the massive-scale network of Topcat, tuning algorithm AROPE for link prediction.}\vspace{-3pt}%
    \label{tbl:topcat}%
    \begin{tabular}{ccccccc}
  \toprule[1.2pt]
  \multirow{2}{*}{\tiny Methods}&\multicolumn{2}{c}{\tiny Trail1}&\multicolumn{2}{c}{\tiny Trail2}&\multicolumn{2}{c}{\tiny Trail3}\\
\cline{2-7}
    &{\tiny AUC} &{\tiny Time(s)}&{\tiny AUC} &{\tiny Time(s)}&{\tiny AUC} &{\tiny Time(s)}\\
    \midrule[0.8pt]
    JITuNE& \textbf{0.7260} & 6206.2 & 0.7307 & 10159.2 & \textbf{0.7418} & 14429.4\\
    AutoNE& 0.7120 & 9054.9 & 0.7284 & 13904.2 & 0.7390 & 19251.6\\
    Random& 0.7208 & 4317.8 & \textbf{0.7389} & 9735.1  & 0.7320 & 13783.2\\
  \bottomrule[1.2pt]
  \end{tabular}
  \end{table}

According to Table~\ref{tbl:flickr}, JITuNE performs the best among the three methods. At the same time, it uses much less time than AutoNE. Moreover, JITuNE has the best performance at the initial trial. This is because the tuning on the synopsis in three runs has enables a good starting point for tuning on the original network.

Furthermore, following the state-of-the-art work of AutoNE, we experiment with the Topcat dataset, which has more than a million nodes. We run the AROPE algorithm and test with the link prediction application. The time for JITuNE to tune the hyperparameters of AROPE can only be used to run three rounds of AROPE on the original graph of Topcat. We summarize the performance results in Table~\ref{tbl:topcat}.

Again, JITuNE achieves the best performance at the end of the trials. The time cost for JITuNE is about the same for random search and much less than that for AutoNE. AutoNE requires a long time to run because it needs to sample sub-networks, to tune the hyperparameters of multiple sub-networks, and to run large matrix computations. The results on Topcat, as well as on Flickr, have validated the capability of JITuNE to work in massive-scale networks.\vspace{-6pt}
\subsection{Stability Analysis: Tuning on Synopsis}

We also test whether the hyperparameter tuning results on network synopses are stable. We repeat the automatic tuning process of JITuNE with AROPE on Arxiv and link prediction for five times. Each of the five tests are run independently, i.e., the sampling process in the hyperparameter space is independent for different tests. We set the time constraint to let JITuNE only validate on the original graph after tuning on the network synopsis, but no further tuning on the original graph. The results are presented in Table~\ref{tbl:stability}. One can notice the highly consistent results of the five tuning processes. The performance results have only a small variance. The stability of JITuNE is thus validated.\vspace{-6pt}

\section{Related Work}%
\label{sec:related}

\textbf{Network Embedding.} Networks in real applications are in massive scale. The common representation of affinity matrix cannot easily be fit into the machine learning settings. Network embedding~\cite{zhang2018network} is proposed to learn the latent representation of nodes for a network. With the embedded representation of a network, the progress made in machine learning can be directly applied to network applications, e.g., link prediction, classification and recommendation.

The algorithms for network embedding can be roughly divided into three categories based on matrix factorization, random walks and deep neural network respectively. The factorization-based methods factorize the adjacency matrix of a network to acquire a low-rank approximation for each node~\cite{arope,m-nmf}. The second category is generally based on random walks over the network. Methods in this category are mainly inspired by word2vec~\cite{word2vec}. The representative methods in this category include DeepWalk~\cite{deepwalk}, LINE~\cite{line} and node2vec~\cite{node2vec}. The third category is based on neural network, e.g., SDNE\cite{sdne}, SDAE\cite{cao2016deep}, and TriDNR\cite{pan2016tri}. Besides, some methods based on convolutional neural network are also proposed recently, such as GCN\cite{gcn}, GraphSage\cite{hamilton2017inductive}, and FastGCN\cite{fastgcn}.
\begin{table}[!t]
    \centering
    \small
    \vspace{-6pt}
    \caption{Stability analysis of hyperparameter tuning on network synopsis, with five independent tuning tests of AROPE on Arxiv for link prediction.}\vspace{-3pt}%
    \label{tbl:stability}%
 \begin{tabular}{cccccc}
\toprule[1.2pt]
  Hyperparameters&{\small Test1} &{\small Test2}&{\small Test3}&{\small Test4}&{\small Test5}\\
  \midrule[0.8pt]
  $w1$& 2.4878 & 2.8302 & 3.4594 & 3.3303 & 3.0960\\
  $w2$& 2.8883 & 3.0268 & 2.9904 & 2.9610 & 2.8455\\
  $w3$& 0.4224 & 0.3171 & 0.6057 & 0.3699 & 0.6430\\
  \midrule[0.2pt]
  $P_G$ & 0.7519 & 0.7526 & 0.7532 & 0.7527 & 0.7524\\
\bottomrule[1.2pt]\vspace{-12pt}
\end{tabular}
  \end{table}

\textbf{Network Reduction.} Due to the inability to analyze massive-scale networks in some cases, approaches have been proposed to reduce the network size such that analysis can be carried out, e.g., community discovery and data summarization. Methods for network reduction can be roughly divided into two categories. The first is by sampling~\cite{graphSampling}, but sampling-based methods cannot capture the global and local information of a network simultaneously. The second is by coarsening~\cite{harp,hgcn,coarsenSpectral}, which can first group nodes according to some rules and then collapse them. Both the hierarchical coarsening process of HARP~\cite{harp} and H-GCN~\cite{hgcn} can effectively capture the local information and the global information such as the disconnected components and the community. The steps after coarsening in HARP and H-GCN can lead to an uncompetitive performance in link prediction tasks, because the global topological information is lost in these steps. However, JITuNE only exploits the coarsening step of these methods and achieve very promising results in hyperparameter tuning.

\textbf{AutoML for the general and network embedding.} Multiple AutoML (automated machine learning) frameworks for hyperparameter tuning are proposed, such as Autotune~\cite{autotune}, Auto-Keras~\cite{autokeras}, and Optuna~\cite{optuna}. AutoNE~\cite{autone} is the related work closest to ours, targeting specifically on NE algorithm tuning. It transfers the knowledge of tuning sub-networks to tuning the whole network. But the sub-networks are randomly sampled such that the sub-network is not guaranteed to be representative of the original network for the applications. The local and the global information of a network might be discarded during sampling. Besides, the meta-learner for transferring knowledge cannot handle categorical hyperparameters. In comparison, JITuNE exploits representative network synopses for tuning and knowledge transfer. Its simple yet powerful framework allows various types of hyperparameters, enabling the users to get a result just in time.\vspace{-12pt}

\section{Conclusion}%
\label{sec:conclude}

In this paper, we study the problem of how to automatically tuning the hyperparameters of network embedding (NE) algorithms in a time acceptable to users. To deal with networks in real applications, we propose JITuNE, a Just-In-Time hyperparameter tuning framework for automatically optimizing towards the best performance of an NE algorithm within a given time. In comparison with the state-of-the-art tuning methods for NE algorithms, JITuNE can achieve obviously better results in a shorter period of time. We conduct extensive experiments with representative NE algorithms of different categories over seven real-world networks. Results demonstrate the effectiveness and efficiency of JITuNE.

\balance

\bibliographystyle{ACM-Reference-Format}
\bibliography{ref}

\typeout{get arXiv to do 4 passes: Label(s) may have changed. Rerun}

\end{document}